\documentclass{article}
\usepackage{iclr2027_conference,times}
\iclrfinalcopy % preprint: use final layout (real authors, no line numbers); header text is overridden below

\usepackage{xcolor}
\definecolor{linkblue}{RGB}{31,78,121}
\usepackage[colorlinks=true,linkcolor=linkblue,citecolor=linkblue,urlcolor=linkblue]{hyperref}
\usepackage{url}
\usepackage{graphicx}
\usepackage{booktabs}
\usepackage{amsmath}
\usepackage{amssymb}
\usepackage{caption}
\usepackage{multirow}
\usepackage{subcaption}
\usepackage{array}
\usepackage{float}     % [H] placement for appendix tables
\usepackage{flafter}   % floats never appear before their place in the text
\usepackage{needspace} % keep headings from being stranded at the foot of a page
\usepackage{placeins}  % \FloatBarrier keeps results floats inside their subsection
\usepackage{titletoc}  % appendix table of contents
\usepackage{etoolbox}
\makeatletter
\renewcommand{\@maketitle}{%
  \vbox{\hsize\textwidth
  {\LARGE\sc \@title\par}
  \vskip 0.3in minus 0.1in
  \begin{center}
  \@author
  \end{center}
  \vskip 0.1in}%
}
\makeatother

\usepackage[most]{tcolorbox}
\usepackage{pgfplots}
\pgfplotsset{compat=1.17}
\usepgfplotslibrary{groupplots}
\newif\ifannotations
\annotationsfalse
\definecolor{annotred}{RGB}{240,40,20}
\definecolor{annotyellow}{RGB}{255,215,0}
\definecolor{annotgreen}{RGB}{30,200,40}
\definecolor{annotnone}{RGB}{255,255,255}
\definecolor{annotnotetext}{RGB}{230,30,20}
\tcbset{annotstyle/.style={enhanced, colback=white, colframe=annot#1, boxrule=1.3pt,
  arc=0pt, boxsep=0pt, left=4pt, right=4pt, top=4pt, bottom=4pt,
  grow to left by=5.3pt, grow to right by=5.3pt, parbox=false}}
\newcommand{\annotnotenode}[1]{\node[anchor=north west, inner sep=0pt, text width=2.7cm,
  font=\footnotesize\sffamily, text=annotnotetext, align=left] at ([xshift=12pt]frame.north east) {#1};}
\ifannotations
  \newtcolorbox{annot}[2][yellow]{annotstyle=#1, breakable, before skip=6pt, after skip=6pt,
    overlay unbroken and first={\annotnotenode{#2}}}
  \newtcolorbox{annotfloat}[2][yellow]{annotstyle=#1, before skip=0pt, after skip=0pt,
    overlay={\annotnotenode{#2}}}
  \newcommand{\placeholder}[1]{{\color{gray}\itshape [#1]}}
\else
  \newenvironment{annot}[2][]{}{}
  
  \newcommand{\placeholder}[1]{}
\fi

\title{How Vision Transformers Represent Position: A Mechanistic Comparison of Absolute and Rotary Positional Encodings}

\author{
{\bf Mahmoud Mannes$^{1}$, Nikhil Maturi$^{2}$, Parva Mehta$^{3}$, Jiwon Jeong$^{4}$, Aravind Singaram Kannappan$^{5}$}\\[4pt]
{\normalsize $^{1}$ESSTHS, $^{2}$Independent Researcher, $^{3}$DJSCE, $^{4}$Lotte Innovate, $^{5}$New York University}\\[6pt]
\texttt{\small mannesmahmoud@gmail.com}
\\ \texttt{\small nikhilmaturi.dev@gmail.com} % Nikhil Maturi -- fill in
\\ \texttt{\small parvamehta26@gmail.com} % Parva Mehta -- fill in
\\ \texttt{\small jiwon-jeong@lotte.net} % Jiwon Jeong -- fill in
\\ \texttt{\small ak12124@nyu.edu} % Aravind Singaram Kannappan -- fill in
}

\begin{document}

\maketitle
\thispagestyle{fancy}
\lhead{Preprint}

\begin{abstract}
Positional embeddings encode the spatial position of each image patch in a Vision Transformer (ViT), allowing the model to distinguish different spatial arrangements of visual content. However, we know little about how the choice of embedding shapes a vision transformer's internal representation. In this work, we show that Absolute Positional Embeddings (APE) and Rotary Positional Embeddings (RoPE) encode spatial position through different internal mechanisms: APE stores position in explicit features, while RoPE builds it inside attention and rebuilds it when it is withheld. Specifically, in APE models, sparse autoencoders recover causal row- and column-specific features, whereas interventions on RoPE reveal distinct rotary components that separately encode row and column information. Additionally, causal interventions selectively disabling RoPE's row or column rotations show that decodable row and column information can be independently removed and, once rotations are restored, partly reconstructed, in part by axis-specialized attention heads. Both mechanisms are causally important for classification: permuting APE's positional features across patches reduces top-1 accuracy by 24--31 points, and ablating the rotations of RoPE's primary rebuilding head reduces it more than ablating those of any other head. Our findings show that the choice of positional encoding fundamentally changes how spatial information is organized and processed inside Vision Transformers. More broadly, our work demonstrates how mechanistic analysis can reveal how models construct, reconstruct, and maintain structured representations, providing a basis for targeted interventions in vision models.
\end{abstract}

\section{Introduction}
\label{sec:intro}

Vision Transformers (ViTs) are now the standard backbone for image understanding, from self-supervised encoders \citep{caron2021emerging,oquab2024dinov2,simeoni2025dinov3} to segmentation models \citep{kirillov2023segment} and the image encoders of vision--language models \citep{radford2021clip}. Many of these applications depend on where objects are in an image, yet because self-attention is invariant to the order of its inputs, a ViT knows a patch's location only through its positional encoding \citep{vaswani2017attention,dosovitskiy2021image}. Recent models have moved from learned absolute positional embeddings (APE) \citep{dosovitskiy2021image} to rotary positional embeddings (RoPE) \citep{su2024roformer,fang2023eva02,heo2024rope,wang2024qwen2vl,simeoni2025dinov3}, a choice made on the basis of accuracy and resolution extrapolation \citep{heo2024rope,fuller2024lookhere}. However, spatial reasoning remains a known bottleneck: vision--language models built on ViT encoders struggle with basic spatial relations, a gap recent work calls their spatial blindspot \citep{alam2026spatial}, and they make little use of spatial cues even when positional encodings are in place \citep{qi2025beyond}. The fixes proposed so far change the encoding itself and are judged only by benchmark scores, without a mechanistic account of how APE and RoPE represent position in ViTs, where that information is held, and how the model uses it.

\begin{figure}[t]
\centering
\includegraphics[width=\textwidth]{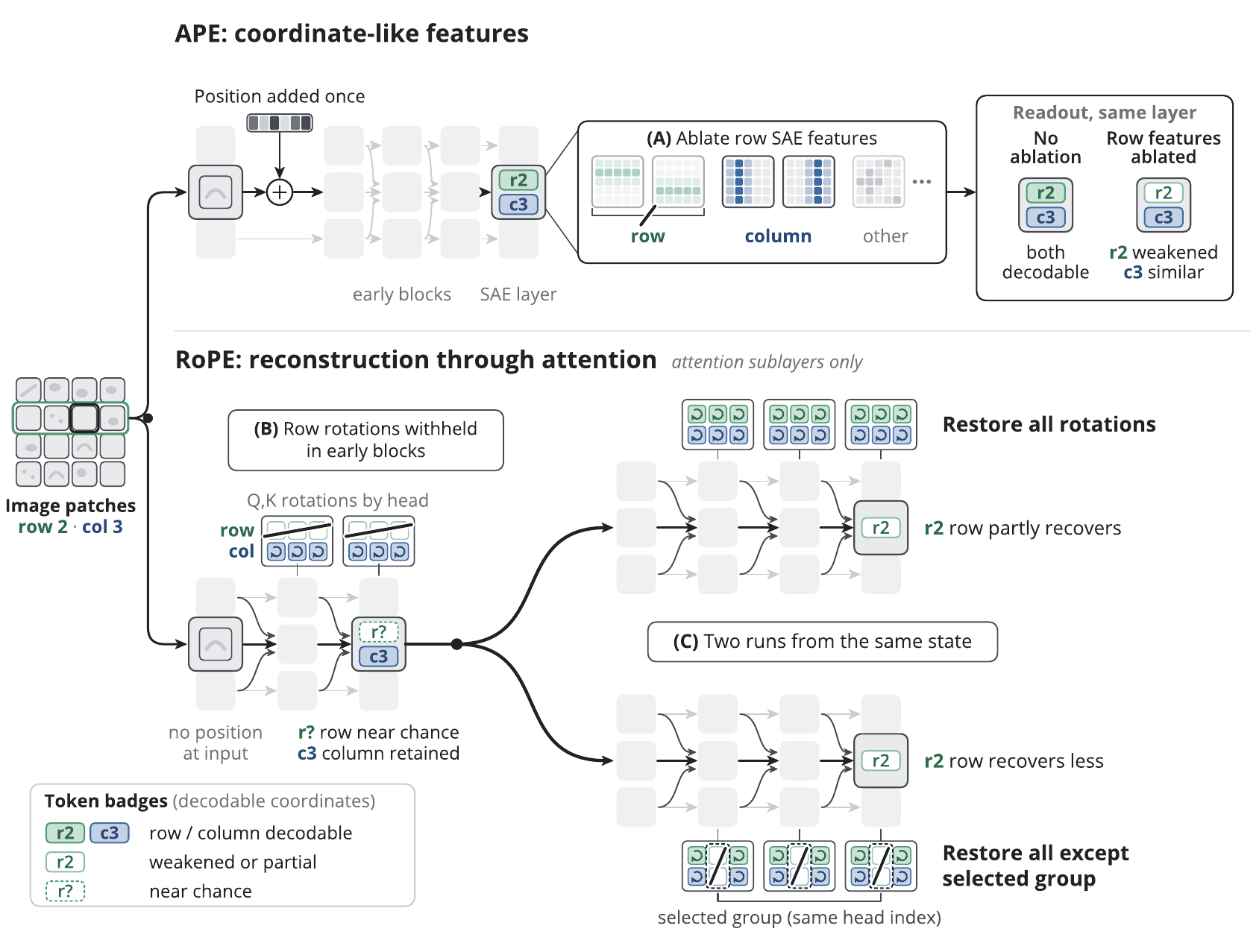}
\caption{\textbf{APE stores position in coordinate-like residual-stream features, whereas RoPE ViTs rebuild position through attention.} Top: APE adds position once at the input, so at the SAE layer a patch token carries decodable row and column coordinates, and (A) ablating the row-selective SAE features weakens the row readout while the column readout stays similar. Bottom: RoPE adds no position at the input and instead rotates queries and keys, per head and separately for rows and columns, in every attention layer. (B) With the row rotations withheld in early blocks, the row coordinate is near chance while the column coordinate is retained. (C) Two runs continue from this state: restoring all rotations lets the row coordinate partly recover, and restoring all except those of a selected head group (the same head index in each block) lets it recover less.}
\label{fig:overview}
\end{figure}

At a technical level, the two encodings differ in how they provide position: APE adds a learned vector to each patch once, before the first block \citep{dosovitskiy2021image}, whereas RoPE leaves the patches unchanged and rotates queries and keys at every layer by angles set by each patch's row and column \citep{su2024roformer,heo2024rope}. Most evidence about position inside trained ViTs comes from linear probes, which show that position can be decoded from a hidden state but not where it came from or whether the model uses it \citep{alain2017understanding,belinkov2022probing}. \citet{mannes2026position} introduced a more direct measure, the spatial similarity--distance correlation (SSDC), which asks whether patches that are close on the image grid have similar hidden states; under random permutation at inference (RPI), which shuffles patch content while position stays attached to each token, it measures structure due to position alone. That work characterized this structure in small ViTs trained from scratch, but did not identify the components that produce it or test whether classification depends on it.

Across four APE and four RoPE ViTs, we show that the two encodings lead the network to represent position through different mechanisms. In APE models, position is stored in the residual stream: a sparse autoencoder (SAE) finds that it is driven by a small set of features, about 20 of 3,840, each tied to a band of rows or columns, which together form a coordinate system over the image grid. Downstream classification depends on these features: permuting them across patches, with the image unchanged, reduces ImageNet top-1 accuracy by 24--31 points. RoPE models contain no such features; position is instead computed inside attention by rotating queries and keys, with separate rotations for rows and columns, so that replacing the row rotations with the identity removes row position while leaving column position intact, and vice versa. Because these rotations are applied again at every block, the model recomputes position throughout the network: when they are withheld from the early blocks and then restored, later blocks rebuild 50--71\% of the lost spatial structure (Figure~\ref{fig:overview}).

These results mechanistically characterize APE and RoPE in vision models, and illustrate how position is represented inside vision transformers and how to intervene on it. For interpretability research, position in an APE model can be read and edited through a few SAE features, whereas in a RoPE model the SAE is unreliable and the rotations themselves are the natural point of intervention; secondarily, for work on spatial reasoning, replacing one encoding with the other changes how position is represented throughout the network, a change that can be measured directly with the tools we use here rather than inferred from downstream benchmark scores.

Our contributions are:

\begin{itemize}
\item \textbf{Mechanistic evidence for APE.} In four pretrained APE ViTs, a sparse autoencoder isolates a small set of row- and column-selective features that tile the image grid, and removing one axis's features removes only that axis.
\item \textbf{A causal decomposition of RoPE.} Replacing RoPE's rotations with the identity removes position exactly and reversibly, by axis, block or head, and shows that each axis is carried by its own rotations.
\item \textbf{Evidence for positional rebuilding.} When rotations are withheld from the early blocks and then restored, later blocks rebuild much of the lost spatial structure, partly through attention heads that each restore mainly one axis.
\item \textbf{Functional evidence.} Both mechanisms matter for ImageNet classification: moving APE's positional features between patches, or ablating the rotations of RoPE's main rebuilding head, degrades accuracy more than matched controls.
\end{itemize}

\section{Related work}
\label{sec:related}

\textbf{Positional encodings.} RoPE was introduced for language models \citep{su2024roformer} and adapted to images by \citet{fang2023eva02} and \citet{heo2024rope}, who found that it matches APE's accuracy at the training resolution and extrapolates better beyond it; three of our checkpoints come from their release. Other work compares encodings by their effect on extrapolation in ViTs \citep{fuller2024lookhere} and length generalisation in language models \citep{kazemnejad2023impact,chi2023dissecting}, or asks how language models learn position without one \citep{haviv2022transformer}. The closest mechanistic study, \citet{barbero2025round}, analyses how Gemma uses the frequencies of its one-dimensional RoPE and tests changes by retraining. We instead compare two encodings side by side in trained ViTs, and edit the row and column rotations directly at inference, with no retraining.

\textbf{Spatial structure in ViTs.} \citet{dosovitskiy2021image} showed that the learned position embeddings of APE ViTs form a row-and-column pattern, and \citet{mannes2026position} used SSDC and RPI to track spatial structure in ViT-S models trained from scratch. Both describe the structure without considering which components produce it or how the model uses it; we answer both questions using pretrained ViTs. Lastly, \citet{darcet2024registers} found artefact tokens in ViTs that added register tokens remove; we test whether such prefix tokens carry position in RoPE models, and find that the image border does instead.

\textbf{Probes, sparse autoencoders and interventions.} Since probes show what can be decoded from a hidden state rather than what the model uses \citep{alain2017understanding,belinkov2022probing}, we identify positional features with sparse autoencoders, which were developed for language models \citep{cunningham2024sparse,bricken2023monosemanticity,templeton2024scaling} and have recently been applied to vision models \citep{lim2025sparse,fel2025archetypal}, and test them with interventions. Causal analyses usually ablate or patch whole heads, layers or residual-stream states \citep{vig2020investigating,elhage2021mathematical,meng2022locating,wang2023interpretability}; for RoPE we edit only the rotations, which removes position while leaving the rest of each head's computation intact.

\section{Methods}
\label{sec:methods}

% iter1: the red-boxed 'Patches, positions, and the two schemes' now lives in Appendix A.1.

\subsection{Models}

% [iter0: YELLOW box]
\begin{annot}[yellow]{Rewritten in iter1}
We study four APE-only ViT-B checkpoints (supervised \citep{dosovitskiy2021image}, DINO \citep{caron2021emerging}, AugReg \citep{steiner2022augreg} and SAM \citep{chen2022vision}), four RoPE-only checkpoints from the \texttt{naver\_in1k} \citep{heo2024rope}, \texttt{sbb\_in1k} \citep{wightman2019timm} and DINOv3 \citep{simeoni2025dinov3} recipes at ViT-B and ViT-S scale, and one hybrid ViT-B that uses both \citep{heo2024rope}. DeiT-III ViT-B \citep{touvron2022deit3}, an APE model trained with the \texttt{naver\_in1k} recipe, is used for one comparison. Supervised ViT-B and naver ViT-B are the primary models; Table~\ref{tab:models} in Appendix~\ref{app:config} lists every checkpoint.
\end{annot}

\subsection{Measuring spatial structure}
\label{sec:setup}

\paragraph{SSDC.} If a model represents where each patch sits, patches that are close on the image grid should have similar hidden states. The spatial similarity--distance correlation (SSDC) of \citet{mannes2026position} measures this: for every pair of patch tokens we take the cosine similarity of their hidden states and compute its Spearman rank correlation with how close the two patches are on the grid (negative Manhattan distance), averaged over images. SSDC is near 1 when nearer patches are consistently more similar and near 0 when similarity ignores grid distance. We measure it on the hidden state entering each of the 12 blocks, read after the block's first LayerNorm, with blocks indexed from 0.

\paragraph{Probes.} On the same hidden states we train linear probes to predict each patch's exact position (196 classes, chance 0.5\%) or its row and column (14 classes each, chance 7.1\%), and, for APE, ridge regressions on the row and column coordinates, reported as $R^2$. Since a probe shows only that position can be read out, we pair every probe result with an intervention; training details are in Appendix~\ref{app:protocol}.

\paragraph{Controlling for image content.} Neighbouring patches often look alike, so SSDC would stay above zero even in a model with no positional information. We therefore compute SSDC and probes under random permutation at inference (RPI), which shuffles the patch tokens after the patch embedding while the positional signal stays attached to its place in the sequence, so any spatial structure that remains comes from position rather than from image content. Removing the positional signal as well gives a floor. For RoPE models, where we compare how much structure returns across windows and checkpoints, we rescale SSDC and probe accuracy so that the floor is 0 and the intact model is 1:
$$
\text{normalised SSDC}_b = \frac{\text{observed}_b-\text{floor}_b}{\text{baseline}_b-\text{floor}_b}.
$$
APE results are unnormalized, since each APE intervention is compared with controls in the same model.

Table~\ref{tab:experiments} in Appendix~\ref{app:experiments} lists every experiment, and Appendix~\ref{app:protocol} gives compute, sample sizes and implementation checks.

% >>> APE  ================================================================
% Section 4 and Figure 1(a) are the APE half.  Replace freely; nothing in
% Sections 5-10 reads a number from here.
\section{Results}
\label{sec:results}

\subsection{Localization: depth profile}
\label{sec:localization}

\begin{figure}[H]
\centering
\includegraphics[width=\textwidth]{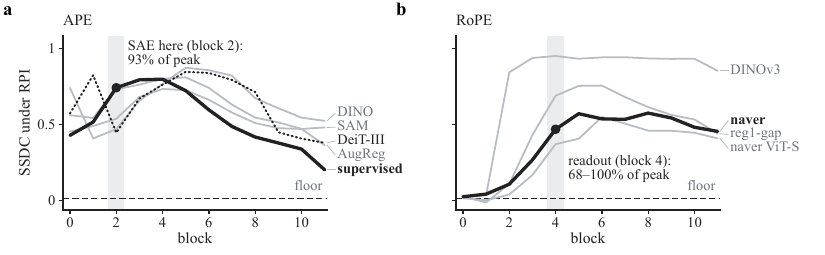}
\caption{\textbf{APE has spatial structure from the input; RoPE builds it over the early blocks.} SSDC under RPI of the residual stream entering each block, for the intact models. Thick lines are the primary models (supervised ViT-B, naver ViT-B) and thin lines the other checkpoints; dashed is the floor with all positional signal removed; dotted is DeiT-III ViT-B (APE), trained with the same recipe as naver ViT-B. Shading marks where later sections act: the SAE at block 2 for APE (Section~\ref{sec:ape_mech}) and the block-4 readout for RoPE (Section~\ref{sec:rope_mech}).}
\label{fig:depth}
\end{figure}

\paragraph{When spatial structure appears.}
SSDC under RPI, measured at every block of the intact models, shows that the two encodings build spatial structure at different depths (Figure~\ref{fig:depth}), and this sets where the later sections intervene. APE adds position at the input, so its structure is present from block 0 and, in supervised ViT-B, reaches 93\% of its peak by block 2. RoPE adds nothing at the input, so SSDC starts at the floor and rises until every RoPE model has 68--100\% of its peak by block 4. We therefore train the APE SAEs on the residual stream entering block 2, where the structure has formed but ten blocks remain in which to see the effect of an ablation, and read the RoPE interventions at block 4, the first block at which all four RoPE models have most of their structure. We chose block 2 on supervised ViT-B, and the other APE models are still building their structure there (54--90\% of their peak), but it is early enough for them too: zeroing their positional features at block 2 leaves spatial structure low, with no recovery at any later block, in all four models (Appendix~\ref{apd:ape_depth_probes}). The difference in depth illustrates the contrasting mechanisms: APE's spatial structure is in place almost from the input, whereas RoPE's is assembled by attention over the first few blocks, a split that activation patching confirms at the level of individual components (Appendix~\ref{sec:results_alignment}).

% ---------------------------------------------------------------------------
% 4.2  = iter0 4.1 (explicit features) + iter0 4.2 (axis-selective ablation)
% ---------------------------------------------------------------------------
\Needspace{4\baselineskip}
\subsection{APE: sparse autoencoder features}
\label{sec:ape_mech}

In each APE model we train an SAE on the residual stream entering block 2 (Section~\ref{sec:localization}) and, for each of the 196 grid positions, keep the most row-selective and the most column-selective feature. Row selectivity is the mean activation of a feature's most active row divided by its mean over the whole grid (1 is uniform, above about 6 is concentrated), and likewise for columns; details are in Appendix~\ref{apd:ape_screen}. We then zero these features and compare against the SAE reconstruction and a random set of the same size.

\begin{figure}[tb]
\centering
\includegraphics[width=\textwidth]{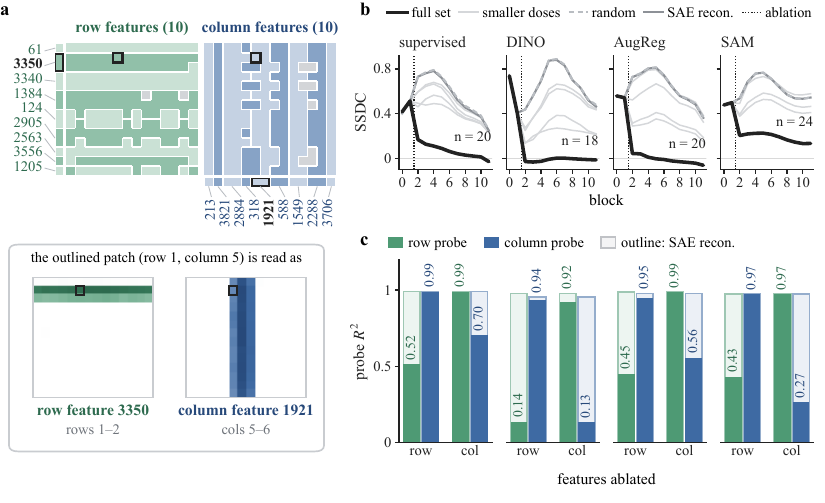}
\caption{\textbf{APE models store position in a small set of coordinate-like SAE features.} We train an SAE on the residual stream entering block 2 and, for each grid position, keep its most row-selective and most column-selective feature. \textbf{(a)} Supervised ViT-B: each patch coloured by the row feature (green) and column feature (blue) that it is assigned. Nine of each tile the grid in ordered bands; grey cells belong to the tenth, which wins only scattered cells. The outlined patch (row 1, column 5) is read by row feature 3350 and column feature 1921, whose mean activation maps are shown below. \textbf{(b)} SSDC by block when the features are zeroed at block 2 (dotted line): the full set of $n$ features (black), smaller sets, a random set of the same size and the SAE reconstruction with nothing zeroed. \textbf{(c)} Row and column probe $R^2$ after zeroing only the row or only the column features; outlines show the SAE reconstruction. Probes fit to shuffled labels read $R^2 \approx 0$.}
\label{fig:ape}
\end{figure}

\paragraph{The features tile the grid.}
The screen finds 20 such features in supervised ViT-B and 18--24 in the other models. In supervised ViT-B, nine of the ten row features and nine of the ten column features each cover a band of one or two adjacent rows or columns, and together they tile the grid in order (Figure~\ref{fig:ape}a). The SAE has found something that looks close to a coordinate system, not a scatter of partly positional features.

\paragraph{Removing them removes the structure.}
\label{sec:ape}
Zeroing the full set drives SSDC to 0.23 or below from block 2 onward in all four models, while zeroing a random set leaves SSDC close to the SAE reconstruction baseline (Figure~\ref{fig:ape}b), even when the random features are matched to the positional ones on firing rate and activation size (Appendix~\ref{apd:ape_controls}). A position probe shows the same: in supervised ViT-B, it recovers a patch's exact position 22.9\% of the time once the features are zeroed, against 99.7\% intact and 99.1\% after zeroing a random set (Appendix~\ref{apd:ape_doseresponse_main}). The SAE does not capture all of the position, since adding its reconstruction error back restores the probe to 57--93\% while SSDC at block 2 stays low (0.07--0.25; Section~\ref{sec:limitations}). Removing fewer features removes less, and most of the drop in SSDC comes with the last few, as the bands in (a) predict: SSDC falls to its floor only once every band is gone. The spatial structure therefore depends on these particular features, not on removing features in general.

\paragraph{Rows and columns are stored separately.}
\label{sec:ape_axis}
In all four models, zeroing only the row features sharply lowers a ridge-regression probe for the row coordinate while the column probe stays at $R^2 \ge 0.92$, and zeroing the column features does the reverse (Figure~\ref{fig:ape}c; Appendix~\ref{apd:ape_axis_crossmodel}). The targeted probe does not reach zero (row $R^2$ 0.51 and column 0.71 in supervised ViT-B), so some position lies outside the features we screened (Section~\ref{sec:limitations}), but what the screen does find splits cleanly by axis. APE therefore stores a patch's row and its column in separate features, and either one can be removed while the other stays intact.

% ---------------------------------------------------------------------------
% 4.3  = iter0 4.3 (rotations, split by axis) + 4.4 (rebuilding) + 4.5 (heads)
% ---------------------------------------------------------------------------
\Needspace{6\baselineskip}
\subsection{RoPE: rotations and rebuilding heads}
\label{sec:rope_mech}

RoPE adds no positional vector to the residual stream; instead, every block rotates each head's queries and keys by angles set by the patch's row and column, so position reaches the residual stream only through attention. An SAE is a poor tool here, since reconstruction alone shifts SSDC by up to $+0.214$ on naver ViT-B (Appendix~\ref{app:sae_rope}), but the rotations themselves can be edited exactly. Each ViT-B head has 32 rotation planes in its queries and keys, half turned by the row and half by the column, and setting a plane's rotation to the identity removes it reversibly, for one axis, a range of blocks or a single head (Appendix~\ref{app:protocol}).

\begin{figure}[tb]
\centering
\includegraphics[width=\textwidth]{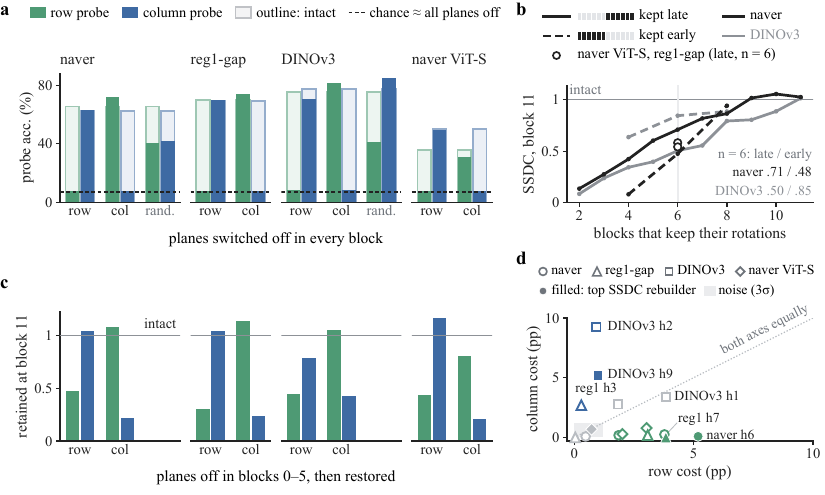}
\caption{\textbf{RoPE position comes from per-axis rotations, is recomputed through depth, and is rebuilt by axis-specialised heads.} Replacing a rotation plane with the identity removes it exactly. \textbf{(a)} Row and column probe accuracy at block 4 with the row or the column planes removed in every block (outline: intact; dashed: chance, which is also where the probes sit with all planes off). The ``rand.'' slot (naver and DINOv3) removes 16 random planes instead, which sends neither axis to chance. \textbf{(b)} Normalised SSDC at block 11 (0: all rotations off, 1: intact) when rotations are kept only in the last $n$ blocks (solid) or only in the first $n$ (dashed); hollow markers show the other two models for the last 6 blocks. \textbf{(c)} Row and column probes at block 11 when one axis's planes are off in blocks 0--5 and restored afterwards (1: intact). \textbf{(d)} Each head's recovery cost: probe points lost on the row ($x$) and column ($y$) probes when that head's rotations stay off in blocks 6--11 during recovery. A head is the same head index in each of those blocks. Colour marks the axis a head favours (grey: no clear preference, or both costs within noise); filled markers are the top rebuilder by SSDC, used in Section~\ref{sec:accuracy}. Grey square: probe noise at three standard deviations; the filled marker inside it (naver ViT-S) has no axis-specific cost.}
\label{fig:rope}
\end{figure}

\paragraph{Each axis has its own rotations.}
\label{sec:planes}
Rotations are the only source of position in these models, and each axis has its own planes. With every plane removed, SSDC falls to 0.011--0.018 and both probes to chance on all four RoPE models. Removing only the row planes sends the row probe to chance and leaves the column probe within 7 points of intact (the gap is largest on DINOv3 and naver ViT-S), removing the column planes does the reverse, and 16 random planes send neither axis to chance (Figure~\ref{fig:rope}a; Appendix~\ref{app:windows}). As in APE, position splits into row and column, but here the split lies in the operation that creates position rather than in features of the residual stream. Rotations encode only offsets between patches, so absolute position must be anchored somewhere. Row and column probes at block 4 are 95--100\% accurate on patches at the image border but 14--56\% in the interior, and edge patches stay at 72--91\% when attention to the class and register tokens is blocked. The prefix tokens do add to the early structure, most of all on DINOv3, whose registers carry four-fifths of its block-4 SSDC, but the absolute position that RoPE models read comes from the edge of the image (Appendix~\ref{app:border}).

\paragraph{Later blocks rebuild position.}
\label{sec:windows}
Because every block applies rotations, later blocks can rebuild position when the early blocks go without it. With all rotations off in blocks 0--5 and restored afterwards, normalised SSDC at block 11 recovers to 0.50--0.71 on all four models, and less comes back as the window grows (Figure~\ref{fig:rope}b). On naver ViT-B, the last six blocks contribute more spatial structure than the first six: keeping rotations only in the last six gives a normalised SSDC of 0.71, against 0.48 when they are kept only in the first six. DINOv3 is the opposite, building early and keeping what it builds (0.85 from the first six blocks against 0.50 from the last six). Removing one axis in blocks 0--5 gives the same picture per axis: by block 11 the missing axis returns to 0.21--0.47 of intact, while the untouched one ends at or above intact on naver ViT-B and reg1-gap and near 0.8 on DINOv3 and naver ViT-S, the two models with cross-axis cost above (Figure~\ref{fig:rope}c; Appendix~\ref{app:windows}). So where APE adds position once at the input, RoPE models build it throughout the network, early or late depending on the model.

\paragraph{Particular heads rebuild each axis.}
\label{sec:heads}
\label{sec:axisheads}
Because rotations are applied separately in each head, we can ask which heads do the rebuilding. We keep one head's rotations off in blocks 6--11 while the rest recover from the 0--5 window, where a head is the same head index in each of those blocks, and rank heads by the SSDC this costs at block 11; the top-ranked head in each model is also the one whose loss costs the most ImageNet accuracy (Section~\ref{sec:accuracy}). Repeating the test with only the row or only the column planes removed shows which axis each head rebuilds (Figure~\ref{fig:rope}d; Appendix~\ref{app:staged}). Most of the top heads rebuild one axis far more than the other, with gaps well above the probe noise of 1.2 points: head 6 of naver ViT-B costs 5.2 points of row recovery and 0.1 of column, head 2 of DINOv3 costs 9.3 points of column recovery and 0.9 of row, and reg1-gap has one head for each axis. The exceptions are one DINOv3 head that costs both axes about equally and the top head of naver ViT-S, which costs neither beyond the noise. Rebuilding is therefore not a general property of the late blocks but the work of particular heads, most of which restore a single axis, so the row--column split that the rotations impose carries through to the heads that recompute position.

\Needspace{6\baselineskip}
\subsection{Comparison: stored once versus recomputed}
\label{sec:compare}

\paragraph{What both encodings share.}
Both encodings represent a row and a column coordinate, and in both the encoding is the only source of it: removing APE's embedding or all of RoPE's rotations leaves SSDC below 0.03 at every block, and removing one axis's SAE features or rotation planes collapses that axis while largely sparing the other (Sections~\ref{sec:ape_axis} and~\ref{sec:planes}).

\paragraph{How they differ.}
If APE stores position and RoPE recomputes it, the two should differ in where spatial structure first appears, whether it takes the form of features, and whether it returns after being removed; all three differences hold. APE's structure is already present entering block 0, whereas RoPE's starts at the floor and is built by attention over the first few blocks (Figure~\ref{fig:depth}); this holds even for DeiT-III and naver ViT-B, which share a training recipe and reach the same top-1 accuracy (Appendix~\ref{app:deit}). APE's structure takes the form of highly selective SAE features in all four models, while the most selective features we find in RoPE models are far less selective (mean selectivity 5.2--9.8 against at most 3.4), although the SAE is also less reliable on RoPE (Appendix~\ref{app:sae_rope}). The most direct test is to remove position partway through the network. Once APE's features are zeroed at block 2, neither SSDC nor a position probe recovers at any later block (Figure~\ref{fig:ape}b; Appendix~\ref{apd:ape_depth_probes}), because nothing downstream adds position again; once RoPE's rotations are withheld from blocks 0--5, later blocks rebuild 50--71\% of the lost structure (Figure~\ref{fig:rope}b), because every block applies the rotations again. APE therefore writes position into the residual stream once and the network carries it forward, whereas RoPE recomputes it at every block.

\paragraph{Both encodings in one model.}
The hybrid ViT-B, trained with both encodings, lets us compare the two within one model. With its rotations off, SSDC follows the intact model through block 6 and then falls to 0.09 by block 11 (intact: 0.43); with its embedding removed, SSDC stays at the floor until block 3 and then climbs to 0.42. Without its rotations the hybrid's position is even more readable than intact (exact-position probe 88.6\% against 85.1\%), yet top-1 falls from 80.5\% to 37.4\%, so in this model the embedding does not stand in for the rotations (Appendix~\ref{app:hybrid}). Even within one network, then, each encoding produces its own mechanism: the embedding provides the spatial structure of the early blocks and the rotations that of the late blocks, so the mechanism follows the encoding even when both share the same weights and training.

% ---------------------------------------------------------------------------
% 4.5  = iter0 Section 5 (does this matter for classification?)
% ---------------------------------------------------------------------------
\Needspace{6\baselineskip}
\subsection{Downstream impact: ImageNet classification}
\label{sec:accuracy}

\begin{figure}[H]
\centering
\includegraphics[width=\textwidth]{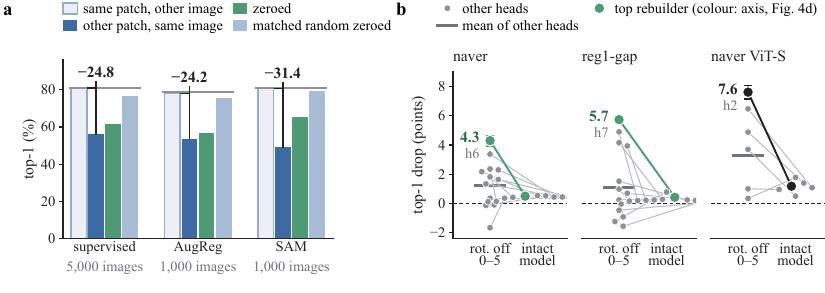}
\caption{\textbf{Both positional mechanisms matter for ImageNet classification.} \textbf{(a)} APE: top-1 after editing the identified features at block 2 with the SAE error added back: swapped with the same patch of another image (outline) or with another patch of the same image (brackets: the gap), zeroed, or replaced by zeroing matched random features. Ticks: intact. DINO has no classification head. \textbf{(b)} RoPE: top-1 drop from removing one head's rotations in blocks 6--11, after rotations were off in blocks 0--5 (left) or in the unmodified model (right). Grey: every other head and their mean; coloured: the top rebuilder, fixed in advance (Section~\ref{sec:rope_mech}) and coloured by favoured axis as in Figure~\ref{fig:rope}d, with its paired standard error. The panels are not on a common scale.}
\label{fig:accuracy}
\end{figure}

We test whether removing APE's positional features, or RoPE's main rebuilding head, costs more ImageNet top-1 accuracy \citep{deng2009imagenet} than removing matched components that carry no position (Figure~\ref{fig:accuracy}).

\paragraph{APE: positional features.}
\label{sec:accuracy_ape}
If the classifier uses APE's positional features, removing them should hurt more than removing other equally active features, and what should matter is the position they report rather than their presence alone. We test both by editing each model's features at block 2 with the SAE's reconstruction error added back, so that only the edited features change. Zeroing them costs 13.9--18.7 top-1 points more than zeroing random features matched on firing rate and activation size (Appendix~\ref{app:cls_ape}), so they matter more than active features that carry no position. Swapping them then shows that the position they encode is what matters. Moving each patch's features to the same patch of another image changes the image they came from but not the position they report, and costs at most 0.7 points; moving them between patches of the same image keeps their image but scrambles their position, and costs 24.8, 24.2 and 31.4 points more on supervised, AugReg and SAM ViT-B (paired standard errors 0.7 and 1.5 on supervised and SAM; Figure~\ref{fig:accuracy}a). On single images, the second swap turns a coffee pot into a water jug while the pixels stay unchanged (Appendix~\ref{app:cls_ape}). Since we chose these features only by where on the grid they fire, never by their effect on the output, their effect on accuracy is an independent finding: ImageNet classification in APE models depends on the position that a small set of residual-stream features encodes.

\paragraph{RoPE: the top rebuilding head.}
\label{sec:accuracy_rope}
In RoPE models the head that does the most rebuilding matters more for classification than any other head. We withhold rotations in blocks 0--5 as in Section~\ref{sec:windows} and then also remove one head's rotations in blocks 6--11, for every head in turn. The top rebuilder was fixed beforehand by its SSDC score (Section~\ref{sec:heads}), so accuracy is an independent check of that ranking. On all three RoPE classifiers it is the costliest head: 4.3, 7.6 and 5.7 points on naver ViT-B, naver ViT-S and reg1-gap, against 3.4, 6.5 and 4.9 for the next head, and 3.1--4.6 points above the mean of the others (paired standard error 0.27 and 0.38 on the two naver models; Figure~\ref{fig:accuracy}b). If the SSDC ranking carried no information, the chosen head would rank first in all three models with probability 1/864. With the early rotations intact the same head costs only 0.4--1.2 points, and giving it rotations for shuffled positions costs about as much as removing them (3.5--6.2 points; Appendix~\ref{app:cls_rope}), so ImageNet classification in RoPE models depends on this head for rebuilding the position that the early blocks lost.

\section{Conclusion and discussion}
\label{sec:limitations}

Both encodings give a vision transformer a row and a column coordinate for each patch, but they store it in different places. An APE model writes position into the residual stream once, in about 20 row- and column-selective features that an SAE can isolate; zeroing these features removes spatial structure from every later block, and moving them between patches lowers top-1 accuracy by 24--31 points. A RoPE model has no comparable features. It computes position inside attention, through rotations that act on each axis separately and are applied again at every block, which lets later blocks rebuild much of the structure lost in the early ones; a few heads do this rebuilding, and most of them restore a single axis. Where a model keeps position is therefore set by its encoding, and this determines how position can be studied and edited: with SAE features in APE models, and with the rotations and the heads that use them in RoPE models.

Three limitations qualify these conclusions. Because we study APE through feature ablation and RoPE through rotation masking, we compare the two mechanisms qualitatively and not by effect size; only one pair of models, DeiT-III and naver ViT-B, shares a training recipe, and all models are ViT-B or ViT-S at 224$\times$224. The APE features also do not account for all decodable position. When the SAE's reconstruction error is added back after the features are zeroed, the exact-position probe at block 2 recovers to 57--93\%, although SSDC stays low and top-1 still falls by 16--22 points, so the features explain the spatial structure and most of its effect on classification but not everything a probe can read. Finally, each head in our RoPE experiments is one head index spanning six blocks, and the attention statistics we examined do not explain what the rebuilding heads compute (Appendix~\ref{app:attn}). Masking single heads in single blocks would address this, and extending the analysis to larger models and higher resolutions, where SSDC becomes less sensitive (Appendix~\ref{apd:ssdc_highres}), would show how far these mechanisms generalise.

\section*{Reproducibility statement}
All checkpoints are public and listed with exact identifiers in Appendix~\ref{app:models}; all images come from the ImageNet-1k validation split. Appendix~\ref{app:protocol} gives sample sizes, seeds, probe and masking protocols, the implementation checks every experiment runs, and the hardware used. Code, SAE weights, the selected feature lists and all result files are available in the anonymous repository linked in Appendix~\ref{app:protocol}.

\bibliographystyle{iclr2027_conference}
\bibliography{refs}

% ===================== APPENDIX DRAFT (iter4, Batch 2) =====================
\providecommand{\confirm}[1]{\textcolor{red}{[#1]}}

\clearpage
\appendix
\section*{Appendix}
\subsection*{Appendix contents}
\startcontents[appendix]
{\footnotesize\hypersetup{linkcolor=black}\printcontents[appendix]{}{1}{\setcounter{tocdepth}{2}}}
\clearpage

% ---------------------------------------------------------------------------
\FloatBarrier
\section{Background, models and data}
\label{app:config}

\FloatBarrier
\subsection{Patches, positions and the two schemes}

A ViT-B/16 at 224$\times$224 splits the image into $14\times14 = 196$ patches, embeds each as a vector, and prepends one or more extra tokens such as the class token; each block then mixes the tokens with self-attention and transforms them with an MLP. APE adds a learned vector $p_{r,c}$ to the patch at row $r$ and column $c$, once, before block 0. After that, position is part of the patch vector, and the network is free to move it, copy it or discard it.

RoPE adds no positional vector to the residual stream. Inside each attention layer it splits the query and key vectors into two-dimensional planes and rotates each plane by an angle proportional to one of the patch's coordinates; a ViT-B RoPE head has 64 dimensions, which pair into 32 planes, 16 turned by the row and 16 by the column. The rotation touches queries and keys but not values, so position enters only through the attention weights, and it enters again at every block. Our experiments rely on this difference: APE's signal is added once and can only be reached through the representation it produces, while RoPE's is an operation repeated at every block that we can replace with the identity wherever we choose.

\FloatBarrier
\subsection{Models and data}
\label{app:models}

Table~\ref{tab:models} lists every checkpoint. All are loaded through \texttt{timm} \citep{wightman2019timm}, except supervised ViT-B, which comes from the HuggingFace hub. Every model is evaluated at 224$\times$224 input, a $14\times14$ patch grid; reg1-gap was trained at 256$\times$256 and runs at 224 through \texttt{timm}'s dynamic image size. ``Prefix'' is the number of non-patch tokens.

For the RoPE models, \texttt{naver\_in1k} is the ImageNet-1k recipe of \citet{heo2024rope}, which follows DeiT-III; \texttt{sbb\_in1k} is the \texttt{timm} ImageNet-1k recipe with a register token and global average pooling; and \texttt{lvd1689m} is DINOv3's self-supervised pretraining. The \texttt{timm} RoPE implementations assign rotation planes to the two axes in opposite orders, and DINOv3 tiles the two components of each plane where the others interleave them, so we derive the plane-to-axis mapping at runtime from each model's frequency table.

All images come from the ImageNet-1k validation split \citep{deng2009imagenet}, streamed from the HuggingFace hub in a fixed order; each experiment uses the first $N$ images of the stream, with $N$ given in Appendix~\ref{app:protocol}.

\begin{table}[!htbp]
\centering
\footnotesize
\setlength{\tabcolsep}{3.5pt}
\caption{Checkpoints. ``IN-21k $\to$ IN-1k'': supervised pretraining on ImageNet-21k, then fine-tuning on ImageNet-1k.}
\label{tab:models}
\begin{tabular}{l>{\scriptsize}lcccl}
\toprule
Short name & \footnotesize Checkpoint & PE & Heads & Prefix & Training \\
\midrule
supervised ViT-B & \texttt{google/vit-base-patch16-224} & APE & 12 & 1 & IN-21k $\to$ IN-1k \\
DINO ViT-B & \texttt{vit\_base\_patch16\_224.dino} & APE & 12 & 1 & DINO, IN-1k \\
AugReg ViT-B & \texttt{vit\_base\_patch16\_224.augreg\_in1k} & APE & 12 & 1 & IN-1k \\
SAM ViT-B & \texttt{vit\_base\_patch16\_224.sam\_in1k} & APE & 12 & 1 & IN-1k \\
DeiT-III ViT-B & \texttt{deit3\_base\_patch16\_224.fb\_in1k} & APE & 12 & 1 & DeiT-III, IN-1k \\
\midrule
naver ViT-B & \texttt{vit\_base\_patch16\_rope\_224.naver\_in1k} & RoPE & 12 & 1 & \texttt{naver\_in1k} \\
naver ViT-S & \texttt{vit\_small\_patch16\_rope\_224.naver\_in1k} & RoPE & 6 & 1 & \texttt{naver\_in1k} \\
reg1-gap ViT-B & \texttt{vit\_base\_patch16\_rope\_reg1\_gap\_256.sbb\_in1k} & RoPE & 12 & 1 & \texttt{sbb\_in1k} \\
DINOv3 ViT-B & \texttt{vit\_base\_patch16\_dinov3.lvd1689m} & RoPE & 12 & 5 & \texttt{lvd1689m} \\
\midrule
hybrid ViT-B & \texttt{vit\_base\_patch16\_rope\_ape\_224.naver\_in1k} & both & 12 & 1 & \texttt{naver\_in1k} \\
\bottomrule
\end{tabular}
\end{table}

\FloatBarrier
\subsection{Terms used in the paper}
\label{app:terms}

\begin{figure}[!htbp]
\centering
\includegraphics[width=\textwidth]{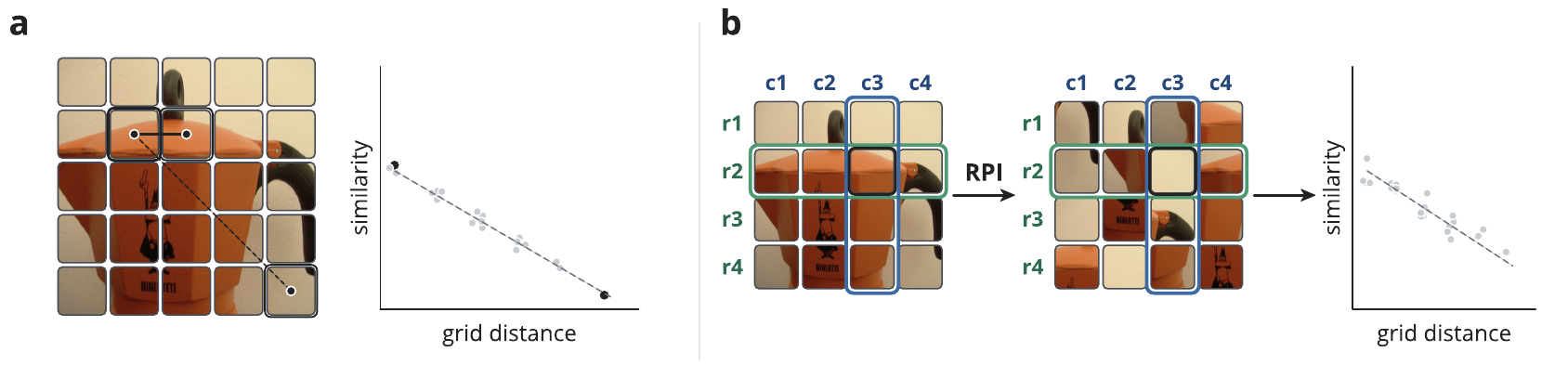}
\caption{\textbf{How SSDC and RPI work}, on a real ImageNet image. \textbf{(a)} For every pair of patches we compare the cosine similarity of their hidden states with their distance on the grid; SSDC is the rank correlation between the two, averaged over images. \textbf{(b)} Under RPI the patch contents are shuffled before block 0 while each sequence index keeps its position, so neighbouring indices no longer hold neighbouring image content. Any SSDC that remains must come from position.}
\label{fig:app_ssdc}
\end{figure}

Table~\ref{tab:terms} collects the terms the paper uses repeatedly, and Figure~\ref{fig:app_ssdc} shows how SSDC and RPI work.

\begin{table}[!htbp]
\centering
\footnotesize
\caption{\textbf{Terms used throughout.}}
\label{tab:terms}
\begin{tabular}{@{}>{\raggedright\arraybackslash}p{2.6cm}>{\raggedright\arraybackslash}p{10.6cm}@{}}
\toprule
Term & Meaning \\
\midrule
Block $b$ & The hidden state entering transformer block $b$ (read after its first LayerNorm), blocks numbered 0--11. The patching experiments of Appendix~\ref{sec:results_alignment} are the one exception: they read the output of block $b$. \\
SSDC & Rank correlation between how similar two patches' hidden states are and how close the patches are on the grid, averaged over patch pairs and images. 1: nearer patches are always more similar; 0: similarity ignores distance. \\
RPI & Random permutation at inference: patch contents are shuffled after the patch embedding, while position stays attached to each sequence index. Structure that survives comes from position, not image content. \\
Floor & The same measurement with all positional signal removed (APE: embedding zeroed; RoPE: every rotation replaced by the identity). \\
Normalised SSDC & $(\text{observed}-\text{floor})/(\text{intact}-\text{floor})$ at the same block: 0 is the floor, 1 the intact model. Used for RoPE only. \\
Probe & Linear classifier for a patch's exact position (196 classes), row or column (14 classes each), trained on hidden states under RPI. \\
Positional features & The row- and column-selective SAE latents kept by the screen of Appendix~\ref{apd:ape_screen}: 18--24 per APE model. \\
Rotation plane & A pair of query/key dimensions that RoPE rotates together, by an angle set by either the row or the column. Masking a plane replaces its rotation with the identity. \\
Window & The blocks in which rotations are removed, e.g.\ ``0--5'': off in blocks 0 to 5, on from block 6. \\
Head & One head index across a range of blocks (e.g.\ head 6 in each of blocks 6--11), not a single head in a single block. \\
Rebuilder & A head whose rotations, when kept off after the 0--5 window, most reduce the SSDC that returns by block 11. \\
Carrier & A head whose rotations, when removed in every block of the unmodified model, most reduce SSDC at block 4. \\
\bottomrule
\end{tabular}
\end{table}

% ---------------------------------------------------------------------------
\FloatBarrier
\section{Experiments at a glance}
\label{app:experiments}

Table~\ref{tab:experiments} lists every experiment with its intervention, controls, readouts and models.

\begin{table}[!htbp]
\centering
\footnotesize
\setlength{\tabcolsep}{3pt}
\caption{Experiments at a glance. ``Floor'': all positional signal removed. ``SAE recon.'': activations passed through the SAE with nothing edited. ``+ error'': the SAE's reconstruction error added back, so only the edited features change.}
\label{tab:experiments}
\begin{tabular}{@{}>{\raggedright\arraybackslash}p{2.2cm}>{\raggedright\arraybackslash}p{3.7cm}>{\raggedright\arraybackslash}p{3.0cm}>{\raggedright\arraybackslash}p{1.8cm}>{\raggedright\arraybackslash}p{2.3cm}@{}}
\toprule
Experiment & Intervention & Controls & Readouts & Models \\
\midrule
Depth profile (\S\ref{sec:localization}) & None & Floor & SSDC per block & all \\
Patching (App.~\ref{sec:results_alignment}) & Copy or zero one attention or MLP output & Clean run, RPI run, misaligned and other-image donors & SSDC, top-1 & sup.\ ViT-B, naver ViT-B \\
APE features (\S\ref{sec:ape_mech}) & Zero SAE features at block 2, by dose or axis & SAE recon., random and matched random features & SSDC, probes & 4 APE \\
APE controls (App.~\ref{apd:ape_controls}) & Zero, mean-ablate or swap features, with error added back & Matched random features; embedding zeroed or permuted & SSDC, probes, top-1 & 4 APE, DeiT-III \\
Rotation planes (\S\ref{sec:planes}) & Identity on all, row or column planes & 16 random planes, floor & SSDC, probes & 4 RoPE \\
Border (App.~\ref{app:border}) & Probe split by distance to border; patch-to-prefix attention blocked & Intact & Probes, SSDC, top-1 & 4 RoPE \\
Windows (\S\ref{sec:windows}) & Rotations off in blocks $0$--$k$, then restored; suffix and middle windows & Intact, floor & Normalised SSDC, probes & 4 RoPE (full sweep on 2) \\
Heads (\S\ref{sec:heads}) & One head's rotations off in blocks 6--11 after the 0--5 window, all planes or one axis & All heads restored & SSDC, probes & 4 RoPE \\
Classification (\S\ref{sec:accuracy}) & Every head's rotations off in 6--11; APE features swapped or zeroed & Other heads; intact model; shuffled rotations; same-patch swap; matched features & Paired top-1 & 3 RoPE, 3 APE \\
\bottomrule
\end{tabular}
\end{table}

% ---------------------------------------------------------------------------
\FloatBarrier
\section{Localization: patching and ablation}
\label{sec:results_alignment}

This appendix gives the patching and zero-ablation experiments behind Section~\ref{sec:localization}, and controls showing that what the patches restore is position rather than image content.

\FloatBarrier
\subsection{Protocol}
\label{apd:patching_protocol}

We run each image twice, once in normal patch order and once under RPI, copy one attention or MLP output from the normal run into the RPI run, and measure recovery as the fraction of the SSDC gap between the two runs that the patch closes. Zero-ablating the same output, with the residual connection kept, tests whether the component is needed at all. Each condition uses supervised ViT-B and naver ViT-B, 512 validation images and two seeds for image order and permutation. Unlike the rest of the paper, these experiments read SSDC from the \emph{output} of block $b$ rather than from the input to block $b$, so ``block $b$'' here corresponds to ``block $b{+}1$'' elsewhere.

\FloatBarrier
\subsection{Results}
\label{apd:patching_full}

\begin{figure}[!htbp]
\centering
\includegraphics[width=\textwidth]{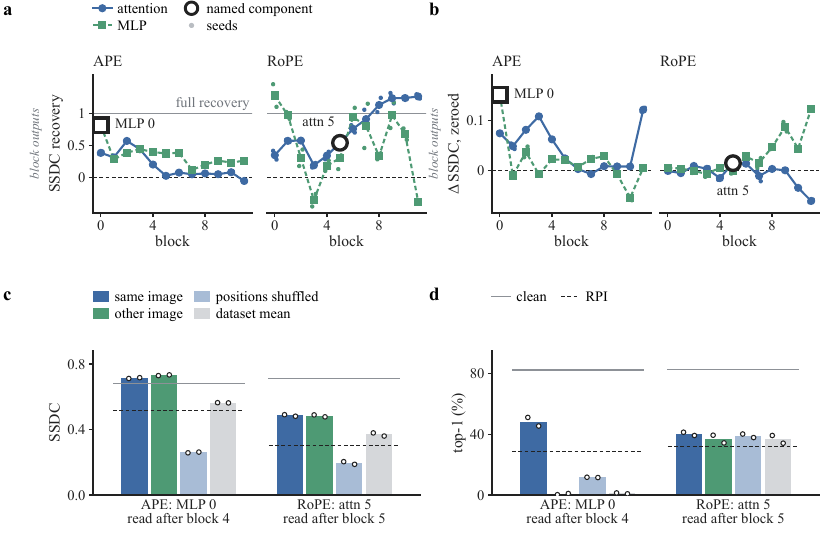}
\caption{\textbf{Patching locates APE's structure in the first MLP and RoPE's in attention over several blocks.} Supervised ViT-B (APE) and naver ViT-B (RoPE), 512 images; small dots are the two seeds. \textbf{(a)} Fraction of the final-block SSDC gap between the normal and RPI runs closed by copying one attention (circles) or MLP (squares) output from the normal run; the components named in the text are ringed. \textbf{(b)} Change in final-block SSDC under RPI when the same output is zeroed. \textbf{(c)} SSDC after patching APE's block-0 MLP (read after block 4) or RoPE's block-5 attention (read after block 5) with activations from the same image, from another image at the same positions, from the same image with positions shuffled, or from the dataset mean; lines mark the clean and RPI runs. \textbf{(d)} Top-1 under the same donors. SSDC is read at block outputs here (Appendix~\ref{apd:patching_protocol}).}
\label{fig:app_patching}
\end{figure}

Figure~\ref{fig:app_patching} gives the full results.

\textbf{APE: the first MLP.} Patching the block-0 MLP alone restores 80--83\% of the final-block SSDC lost under RPI in the two runs, and more than all of it at blocks 4 and 5 (121\% and 105\%). Patching later components restores accuracy but little structure: block-11 attention restores 97--100\% of the lost accuracy.

\textbf{RoPE: attention over several blocks.} Patching block-5 attention restores 45\% of the SSDC lost after that block, and zero-ablating attention in block 2, 3 or 4 lowers block-4 SSDC by 0.079--0.088 in both runs while changing accuracy on unpermuted images by less than 0.5 points.

\textbf{How long attention keeps adding structure.} Attention adds structure in more blocks in RoPE than in APE: under RPI it raises residual SSDC in blocks 0--7 in RoPE (most in blocks 4 and 5, by about 0.12 each) but only in blocks 0--2 in APE (most in block 1, by 0.222), and lowers it after block 3.

\FloatBarrier
\subsection{Alignment controls}
\label{apd:alignment}

Under RPI the patch is copied back into the permuted sequence at matching positions, so a patch could restore SSDC by restoring image content rather than position. We test this at APE's block-0 MLP and RoPE's block-5 attention with four donors (Figure~\ref{fig:app_patching}c,d). Taking the donor activations from a different image, at the same positions, leaves the recovered SSDC unchanged (0.715 to 0.732 in APE, 0.486 to 0.484 in RoPE), while shuffling the donor's positions within the same image drops it to 0.260 and 0.196. A dataset-mean donor, which carries no image content at all, still restores 0.56 in APE. In APE the different-image donor also drops top-1 from 48.3\% to 0.6\%, so the recovered structure is positional: it survives a change of image that destroys classification, and disappears when positions are misaligned.

% ---------------------------------------------------------------------------
\FloatBarrier
\section{Sparse autoencoder analyses}
\label{app:sae}

This appendix supports Section~\ref{sec:ape_mech}: the SAE configuration, the feature screen, the dose-response on all four APE checkpoints, the controls for the ablation, the axis test, and why the SAE is not a reliable instrument on RoPE models.

\FloatBarrier
\subsection{Training and configuration}
\label{apd:sae_config}

Each APE model has one SAE, trained on its residual stream entering block 2 (Table~\ref{tab:sae_config}).

\begin{table}[!htbp]
\centering
\footnotesize
\caption{\textbf{SAE configuration.} One SAE per APE checkpoint, all with the same architecture and training setup; the learning rate and the held-out metrics differ by model.}
\label{tab:sae_config}
\begin{tabular}{@{}l>{\raggedright\arraybackslash}p{9.4cm}@{}}
\toprule
Site & residual stream entering block 2, before LayerNorm \\
Architecture & TopK \citep{gao2025scaling}, $k=64$, with the AuxK loss for dead latents and a unit-norm decoder \\
Latents & 3{,}840 ($5\times$ the width of 768) \\
Training data & all 197 tokens (class token included) of 20{,}000 ImageNet images, scaled so that $\mathbb{E}\|x\| = \sqrt{768}$; 5\% of images held out \\
Optimisation & AdamW \citep{loshchilov2019decoupled}, batch 512 tokens, 2 epochs, 5\% linear warm-up \\
\bottomrule
\end{tabular}

\vspace{6pt}
\begin{tabular}{lccc}
\toprule
Model & Learning rate & Held-out FVU & Dead latents \\
\midrule
supervised ViT-B & $4\times10^{-4}$ & 0.142 & 3.3\% \\
DINO ViT-B       & $2\times10^{-4}$ & 0.139 & 0.0\% \\
AugReg ViT-B     & $4\times10^{-4}$ & 0.126 & 5.7\% \\
SAM ViT-B        & $1\times10^{-4}$ & 0.114 & 0.0\% \\
\bottomrule
\end{tabular}
\end{table}

\FloatBarrier
\subsection{Feature screen}
\label{apd:ape_screen}

\begin{figure}[!htbp]
\centering
\includegraphics[width=\textwidth]{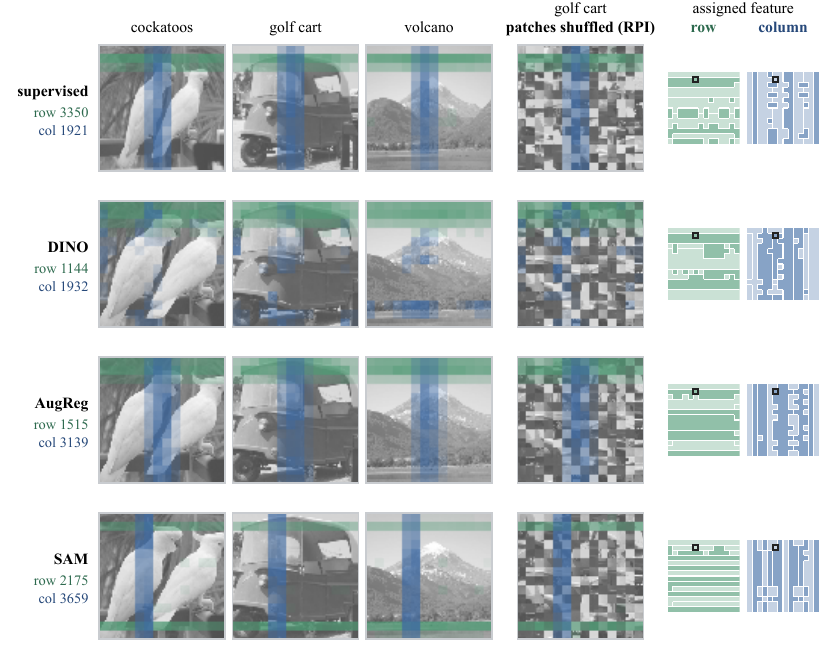}
\caption{\textbf{Positional features respond to position, not content.} For each APE model, the row feature (green) and column feature (blue) that the screen assigns to the patch at row 1, column 5, overlaid on three ImageNet validation images (residual stream entering block 2). Each feature lights the same band on every image, and with the image's patches shuffled (fourth column) the content moves but the bands stay put. Right: the row and column feature assigned to every position; the outlined cell is row 1, column 5. DINO's column feature is shown to scale, fainter than the others.}
\label{fig:app_real_features}
\end{figure}

For each of the 196 grid positions (excluding the class token) we shortlist SAE features by their mean activation at that position minus their mean activation elsewhere, score each candidate's $14\times14$ mean-activation map for row and column selectivity, and keep the best row feature and the best column feature. Row selectivity is the mean activation of a feature's most active row divided by its mean over the whole grid, and likewise for columns. Many positions share a feature, so the screen keeps 20 features in supervised ViT-B, 18 in DINO, 20 in AugReg and 24 in SAM. Each feature counts as a row or a column feature by whichever selectivity is larger. On three of the four checkpoints a few features win positions on both axes, so the row and column subsets overlap slightly and their sizes sum to more than the full set.

Figure~\ref{fig:app_real_features} shows what these features do on real images. For each model we take the row and column features the screen assigns to one patch (row 1, column 5, as in Figure~\ref{fig:ape}a) and plot their activation on every patch of three ImageNet validation images. Each feature lights the same band on the cockatoos, the golf cart and the volcano, and shuffling the image's patches moves the content but not the band, so the features respond to where a patch is, not to what it shows. Two features are less clean: SAM's row feature has a second band at row 12, and DINO's column feature is weak and diffuse.

\FloatBarrier
\subsection{Dose-response}
\label{apd:ape_doseresponse_main}

We zero features in selectivity order at increasing doses and train a linear probe for exact position on the residual stream under RPI (Table~\ref{tab:ape_dose}). On every checkpoint, the SAE round trip and a random set of the same size leave the probe near its intact value, and accuracy falls steadily as more of the selected features are removed. Most of the probe's drop in supervised ViT-B comes between 4 and 8 features, while most of the SSDC drop comes with the last few (Figure~\ref{fig:ape}b): the probe loses precision as soon as a few bands go, but SSDC, which only needs nearby patches to be more similar than distant ones, holds until almost every band is gone.

\begin{table}[!htbp]
\centering
\small
\caption{\textbf{Exact-position probe accuracy (\%) at block 2 as positional features are zeroed.} Chance is 0.5\%. Supervised ViT-B uses 10{,}000 images, the others 1{,}000. The random control zeroes as many random features as the full set. ``---'': dose not run.}
\label{tab:ape_dose}
\begin{tabular}{lcccc}
\toprule
& supervised & DINO & AugReg & SAM \\
\midrule
Intact (no SAE)       & 99.99 & 95.9 & 99.94 & 100 \\
SAE reconstruction    & 99.7  & 82.1$^{\ast}$ & 99.96 & 100 \\
Random features zeroed & 99.1 & 83.4 & 99.8  & 100 \\
\midrule
4 positional features  & 95.9 & 58.7 & 95.0 & --- \\
8                      & 60.5 & 28.4 & 64.7 & --- \\
12                     & 44.0 & 18.2 & 53.7 & --- \\
16                     & 33.2 & 6.6  & 36.8 & --- \\
Full set (20 / 18 / 20 / 24) & 22.9 & 4.3 & 29.3 & 11.3 \\
\bottomrule
\end{tabular}

\vspace{2pt}
{\footnotesize $^{\ast}$A rerun with a validation-selected epoch (Table~\ref{tab:ape_controls}) finds a smaller reconstruction cost on DINO (99.9\% to 96.8\%). SAM's intermediate doses are 7 and 14 features (77.5\% and 27.5\%).}
\end{table}

\FloatBarrier
\subsection{Probes through depth}
\label{apd:ape_depth_probes}

Table~\ref{tab:ape_depth_probes} follows exact-position probe accuracy through all twelve blocks after the features are zeroed at block 2. Once the full set is gone, accuracy stays low all the way to block 11 on every model and never rises by more than 1.5 points, while a random set of the same size leaves every block at the reconstruction level.

\begin{table}[!htbp]
\centering
\scriptsize
\setlength{\tabcolsep}{2.6pt}
\caption{\textbf{Exact-position probe accuracy (\%) at every block after zeroing features at block 2.} Linear probe for a patch's grid position (196 classes, chance 0.5\%) on the residual stream entering each block, under RPI, 1{,}000 images per model. Blocks 0--1 come before the edit. The random control zeroes 20 random features (24 on SAM). Probes: learning rate $5\times10^{-3}$, weight decay $10^{-4}$, batch 512, 20 passes, 10\% of tokens held out; we report the best held-out accuracy over passes.}
\label{tab:ape_depth_probes}
\begin{tabular}{llcccccccccccc}
\toprule
& & \multicolumn{12}{c}{Block} \\
\cmidrule(lr){3-14}
Model & Condition & 0 & 1 & 2 & 3 & 4 & 5 & 6 & 7 & 8 & 9 & 10 & 11 \\
\midrule
\multirow{4}{*}{supervised} & intact (no SAE) & 100.0 & 100.0 & 100.0 & 98.9 & 95.1 & 71.3 & 44.7 & 32.1 & 26.8 & 21.5 & 15.9 & 8.1 \\
 & SAE reconstruction & 100.0 & 100.0 & 100.0 & 98.6 & 92.5 & 66.6 & 40.7 & 30.2 & 24.0 & 20.0 & 15.1 & 7.5 \\
 & all 20 positional zeroed & 100.0 & 100.0 & 21.5 & 20.5 & 18.5 & 14.4 & 9.7 & 7.3 & 6.6 & 5.8 & 5.0 & 3.2 \\
 & 20 random zeroed & 100.0 & 100.0 & 100.0 & 98.6 & 92.4 & 66.6 & 41.1 & 29.8 & 24.5 & 20.0 & 14.6 & 7.2 \\
\midrule
\multirow{4}{*}{DINO} & intact (no SAE) & 69.2 & 93.7 & 96.6 & 95.0 & 90.1 & 84.4 & 79.7 & 75.7 & 72.9 & 70.4 & 69.6 & 68.8 \\
 & SAE reconstruction & 69.2 & 93.7 & 85.8 & 90.8 & 89.2 & 87.3 & 84.8 & 82.8 & 80.3 & 77.9 & 77.4 & 77.3 \\
 & all 18 positional zeroed & 69.2 & 93.7 & 3.8 & 3.9 & 3.8 & 3.9 & 4.0 & 4.2 & 4.3 & 4.7 & 5.1 & 5.3 \\
 & 20 random zeroed & 69.2 & 93.7 & 86.0 & 90.8 & 89.3 & 87.4 & 84.8 & 82.9 & 80.3 & 77.8 & 77.5 & 77.2 \\
\midrule
\multirow{4}{*}{AugReg} & intact (no SAE) & 100.0 & 100.0 & 99.9 & 99.8 & 99.0 & 93.5 & 86.3 & 76.5 & 66.9 & 57.1 & 47.1 & 35.3 \\
 & SAE reconstruction & 100.0 & 100.0 & 99.8 & 99.2 & 98.0 & 89.5 & 80.4 & 72.1 & 61.9 & 52.8 & 43.9 & 33.0 \\
 & all 20 positional zeroed & 100.0 & 100.0 & 28.4 & 24.0 & 20.5 & 17.3 & 13.9 & 11.4 & 9.0 & 7.3 & 5.9 & 4.1 \\
 & 20 random zeroed & 100.0 & 100.0 & 99.8 & 99.3 & 97.8 & 89.2 & 80.6 & 70.9 & 61.7 & 52.7 & 44.1 & 32.9 \\
\midrule
\multirow{4}{*}{SAM} & intact (no SAE) & 99.9 & 100.0 & 100.0 & 100.0 & 100.0 & 98.6 & 97.5 & 96.1 & 94.4 & 93.2 & 92.1 & 91.3 \\
 & SAE reconstruction & 99.9 & 100.0 & 100.0 & 100.0 & 99.9 & 97.8 & 96.4 & 94.9 & 92.5 & 91.4 & 89.9 & 88.9 \\
 & all 24 positional zeroed & 99.9 & 100.0 & 11.3 & 12.2 & 12.8 & 12.5 & 12.2 & 12.3 & 12.3 & 12.3 & 11.8 & 11.7 \\
 & 24 random zeroed & 99.9 & 100.0 & 100.0 & 100.0 & 99.9 & 97.7 & 96.3 & 94.8 & 92.4 & 91.2 & 89.6 & 88.8 \\
\bottomrule
\end{tabular}
\end{table}

\FloatBarrier
\subsection{Controls: matched features, the SAE error and swaps}
\label{apd:ape_controls}

\begin{figure}[!htbp]
\centering
\includegraphics[width=\textwidth]{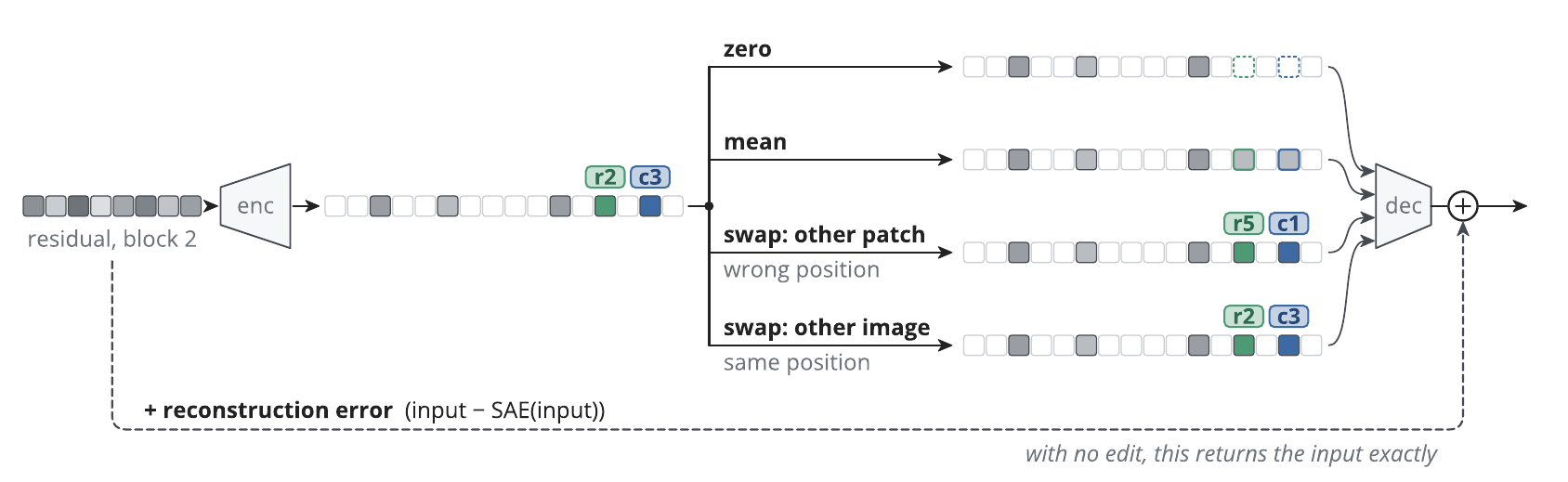}
\caption{\textbf{The edits applied to the positional features at block 2.} Each patch's residual stream is encoded by the SAE; the positional features (coloured) are then zeroed, replaced by their mean, swapped with those of another patch in the same image, or swapped with those of the same patch in another image, and decoded. Adding the reconstruction error back (bottom path) makes the unedited model exactly the intact one, so only the edited features change.}
\label{fig:app_ape_edits}
\end{figure}

\begin{figure}[!htbp]
\centering
\includegraphics[width=\textwidth]{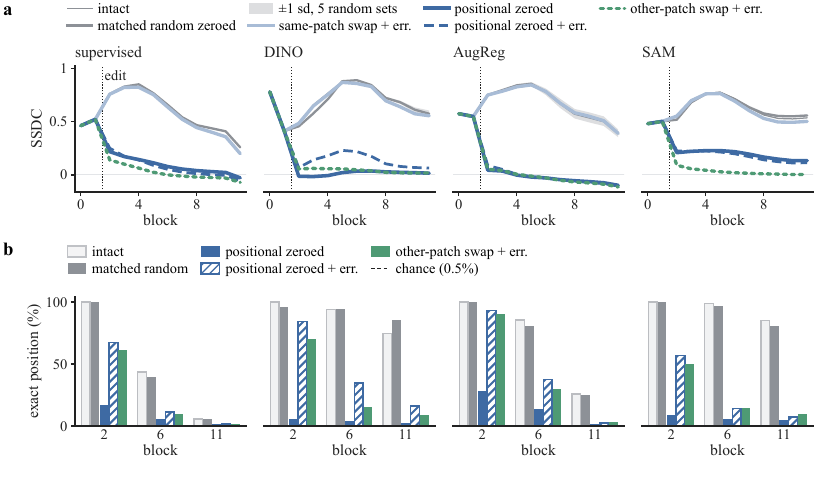}
\caption{\textbf{Only edits that remove or scramble the positional features remove spatial structure, and it does not come back.} \textbf{(a)} SSDC under RPI entering each block after an edit at block 2 (dotted line), for the four APE checkpoints; the band is $\pm$1 standard deviation over five matched random sets. \textbf{(b)} Exact-position probe accuracy at blocks 2, 6 and 11 (chance 0.5\%). ``+ err.'': with the SAE reconstruction error added back. 1{,}000 images.}
\label{fig:app_ape_controls}
\end{figure}

Comparing against a random set leaves three questions open: whether it is a fair comparison, since most SAE latents fire rarely; whether the result depends on what the SAE fails to reconstruct; and whether zeroing hurts the model for reasons unrelated to position. Figure~\ref{fig:app_ape_edits} shows the edits we use to answer them, and Table~\ref{tab:ape_controls} and Figure~\ref{fig:app_ape_controls} give the results.

\textbf{Matched random features.} The positional features fire on 14--31\% of tokens, against a dictionary median below 0.5\%, so a uniformly drawn random set is mostly silent. We therefore draw five random sets matched to the positional set on firing rate and activation size. Zeroing them leaves SSDC and the exact-position probe at the reconstruction level on every model, although it costs 2--5 top-1 points beyond the reconstruction (Appendix~\ref{app:cls_ape}), so they are active features that carry no position.

\textbf{The SAE error.} With the reconstruction error added back, the unedited model is exactly the intact one (identical SSDC and top-1), so only the edited features change. Zeroing the positional features then still leaves block-2 SSDC at 0.07--0.25, but the exact-position probe recovers to 57--93\%. Linearly decodable position therefore also sits in what the SAE does not reconstruct, while the similarity structure that SSDC measures is carried by the features.

\textbf{Mean ablation and swaps.} Replacing the features with their mean has the same effect as zeroing them. Swapping them between images at the same patch changes nothing measurable, while swapping them between patches of the same image, so that each patch reports another patch's position, removes SSDC and most of the probe accuracy. What matters is the position the features report, not how strongly they fire.

\textbf{No rebuilding.} After the features are zeroed at block 2, the exact-position probe stays low through block 11 on every model (Appendix~\ref{apd:ape_depth_probes}), and SSDC does not recover. The one partial exception is DINO with the error added back, whose SSDC recovers to 0.22 at block 6 (intact: 0.86) before falling again.

\begin{table}[!htbp]
\centering
\footnotesize
\setlength{\tabcolsep}{4pt}
\caption{\textbf{APE controls at the intervention site.} SSDC and exact-position probe accuracy (\%) at block 2 after each edit, 1{,}000 images. Matched random: SSDC is the mean of five draws, the probe uses the first draw. The embedding rows remove or permute APE's added embedding before block 0 and use no SAE.}
\label{tab:ape_controls}
\begin{tabular}{lcccccccc}
\toprule
& \multicolumn{2}{c}{supervised} & \multicolumn{2}{c}{DINO} & \multicolumn{2}{c}{AugReg} & \multicolumn{2}{c}{SAM} \\
\cmidrule(lr){2-3}\cmidrule(lr){4-5}\cmidrule(lr){6-7}\cmidrule(lr){8-9}
Edit at block 2 & SSDC & probe & SSDC & probe & SSDC & probe & SSDC & probe \\
\midrule
Intact & 0.76 & 100.0 & 0.47 & 99.9 & 0.75 & 100.0 & 0.55 & 100.0 \\
SAE reconstruction & 0.75 & 100.0 & 0.44 & 96.8 & 0.74 & 99.9 & 0.54 & 100.0 \\
Positional zeroed & 0.22 & 16.6 & $-$0.01 & 5.9 & 0.04 & 28.4 & 0.21 & 8.8 \\
Uniform random zeroed & 0.75 & 100.0 & 0.44 & 97.3 & 0.75 & 99.9 & 0.54 & 100.0 \\
Matched random zeroed & 0.76 & 100.0 & 0.45 & 96.1 & 0.75 & 99.9 & 0.52 & 100.0 \\
\midrule
\multicolumn{9}{l}{\emph{With the SAE error added back}} \\
Positional zeroed & 0.25 & 67.2 & 0.07 & 84.4 & 0.09 & 93.2 & 0.23 & 57.0 \\
Positional mean-ablated & 0.24 & --- & 0.09 & --- & 0.10 & --- & 0.15 & --- \\
Swapped, same patch of another image & 0.76 & 99.9 & 0.48 & 97.4 & 0.75 & 99.9 & 0.55 & 100.0 \\
Swapped, other patch of the same image & 0.14 & 61.3 & 0.06 & 70.4 & 0.05 & 90.0 & 0.09 & 50.2 \\
Matched random zeroed & 0.76 & --- & 0.48 & --- & 0.75 & --- & 0.54 & --- \\
\midrule
Embedding zeroed & 0.01 & 0.6 & 0.01 & 0.5 & 0.01 & 0.5 & 0.01 & 0.6 \\
Embedding permuted & 0.01 & 100.0 & 0.01 & 99.9 & 0.01 & 100.0 & 0.00 & 100.0 \\
\bottomrule
\end{tabular}
\end{table}

Permuting the added embedding across patches leaves exact position fully decodable, since each index still receives a fixed vector, but removes SSDC, because neighbouring indices no longer receive similar vectors. SSDC therefore measures whether position is organised like the grid, which a probe for exact position does not.

\FloatBarrier
\subsection{Axis separation}
\label{apd:ape_axis_crossmodel}

Figure~\ref{fig:ape}c summarises the axis test; Table~\ref{tab:ape_axis_crossmodel} gives the numbers. The probes are ridge regressions on the normalised row and column coordinates, reported as test $R^2$ on a 600/200/200 split by image, with a control fit to shuffled labels. DINO, AugReg and SAM use 1{,}000 images.

\begin{table}[!htbp]
\centering
\small
\caption{\textbf{Zeroing one axis's features removes that axis.} Test $R^2$ of ridge probes for the row and column coordinates at block 2. ``Own axis off'' zeroes the row features for the row probe and the column features for the column probe. Bold: the probe whose own features were removed. Probes fit to shuffled labels read $R^2 \approx 0$ throughout.}
\label{tab:ape_axis_crossmodel}
\begin{tabular}{lcccc}
\toprule
& Intact & SAE recon. & Random & Own axis off \\
\midrule
\multicolumn{5}{l}{\emph{Row probe}} \\
supervised & 0.992 & 0.990 & 0.989 & \textbf{0.515} \\
DINO       & 0.990 & 0.976 & 0.976 & \textbf{0.137} \\
AugReg     & 0.991 & 0.987 & 0.987 & \textbf{0.451} \\
SAM        & 0.991 & 0.972 & 0.972 & \textbf{0.432} \\
\midrule
\multicolumn{5}{l}{\emph{Column probe}} \\
supervised & 0.988 & 0.990 & 0.989 & \textbf{0.705} \\
DINO       & 0.989 & 0.954 & 0.953 & \textbf{0.134} \\
AugReg     & 0.974 & 0.977 & 0.978 & \textbf{0.557} \\
SAM        & 0.981 & 0.973 & 0.973 & \textbf{0.266} \\
\bottomrule
\end{tabular}
\end{table}

Zeroing one axis's features drives that axis's probe to $R^2 \le 0.56$ while the other stays above 0.92, and neither the reconstruction nor the random control moves either probe by more than 0.03.

\textbf{Cost to the other axis.} The cost to the other axis is at most 0.001 on supervised ViT-B and SAM, 0.031 on AugReg, and 0.017 and 0.055 on DINO, the largest of the four; the largest cross-axis cost among the RoPE models is also on a DINO-trained model (Section~\ref{sec:planes}), which points to the training objective rather than the encoding.

\textbf{Both axes together.} Zeroing both subsets together in supervised ViT-B leaves the row probe at 0.514 and the column probe at 0.715, the same as zeroing each alone, so the two subsets act independently. No probe reaches zero, so some per-axis information lies outside the screened features (Appendix~\ref{apd:ape_controls}). The supervised model loses more on the row axis than on the column axis ($\Delta R^2$ 0.475 against 0.285); since the screen and the SAE may resolve the two axes unevenly, we do not read this as a property of the model.

\FloatBarrier
\subsection{RoPE: SAE reconstruction is not a reliable intervention}
\label{app:sae_rope}

The same SAE approach does not transfer to RoPE models, for two reasons. First, the reconstruction alone moves SSDC too much: passing naver ViT-B's residual stream through its SAE and back raises SSDC by up to 0.214 before any feature is removed, against 0.062 and 0.065 on the two APE models measured the same way. Second, on DINOv3 no dose moves SSDC enough: zeroing 22 features, the largest dose we tried, lowers SSDC only from 0.948 to 0.863 at the intervention site, well short of the 30\% drop we set in advance as the bar for a usable edit.

The screen also finds nothing coordinate-like in RoPE models. At block 2, the screened APE features have a mean row and column selectivity of 5.2--9.8 (supervised 7.2 and 7.9, DINO 5.2 and 5.7, AugReg 6.6 and 7.5, SAM 9.8 and 9.1). On naver ViT-B the most selective features reach at most 3.4 at blocks 2, 4, 6, 8 and 10 (row 3.1, 3.4, 3.0, 3.2, 1.8; column 2.2, 3.3, 3.0, 1.9, 1.3), and on DINOv3 at most 1.8 at block 4. We therefore intervene on RoPE's rotations directly, which can be edited exactly by axis, block and head, with no reconstruction step.

% ---------------------------------------------------------------------------
\FloatBarrier
\section{Rotation masking}
\label{app:windows}

\begin{figure}[!htbp]
\centering
\includegraphics[width=\textwidth]{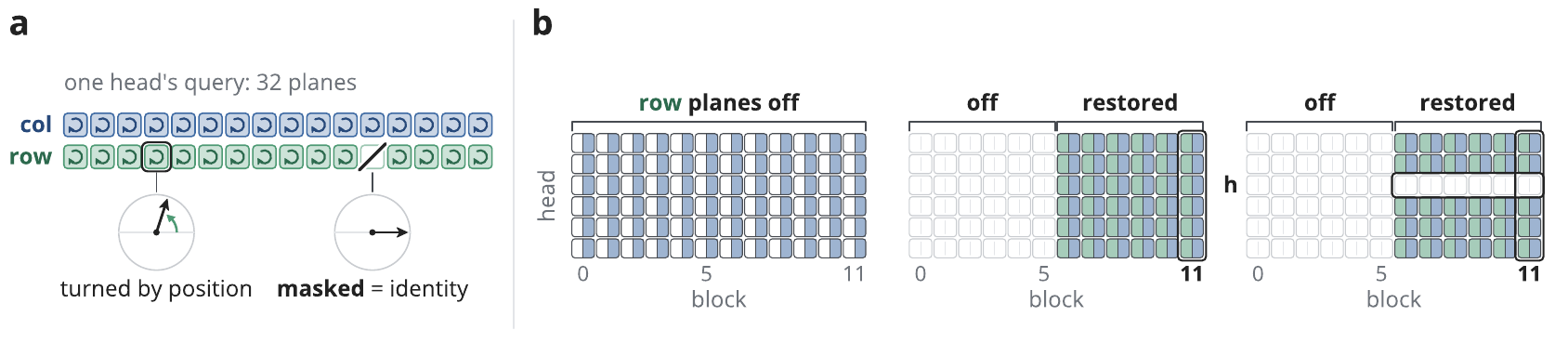}
\caption{\textbf{Rotation masking.} \textbf{(a)} A RoPE head rotates 32 planes of its queries and keys, 16 by the patch's column and 16 by its row; masking replaces a plane's rotation with the identity. \textbf{(b)} The three scopes we use, shown on a grid of heads $\times$ blocks: one axis's planes off in every block (left); every rotation off in blocks 0--5 and restored from block 6, with SSDC and probes read at block 11 (middle); and the same, with one head index $h$ also kept off in blocks 6--11 (right).}
\label{fig:app_masking}
\end{figure}

This appendix supports Sections~\ref{sec:planes} and~\ref{sec:windows}: the axis-removal numbers behind Figure~\ref{fig:rope}a, where RoPE models anchor absolute position, and every masking window behind Figure~\ref{fig:rope}b,c. Figure~\ref{fig:app_masking} shows what the masking does.

\FloatBarrier
\subsection{Removing one axis}

\begin{table}[!htbp]
\centering
\small
\setlength{\tabcolsep}{4pt}
\caption{\textbf{Removing one axis's rotation planes in every block}, read at block 4 (1{,}000 images). Bold: the probe whose own planes were removed; chance is 7.1\%. With every plane off (the floor), SSDC is 0.011--0.018 and both probes 7.5--8.1\% on all four models.}
\label{tab:axis}
\begin{tabular}{lccc@{\hskip 10pt}ccc@{\hskip 10pt}ccc}
\toprule
& \multicolumn{3}{c}{SSDC} & \multicolumn{3}{c}{Row probe (\%)} & \multicolumn{3}{c}{Column probe (\%)} \\
\cmidrule(lr){2-4}\cmidrule(lr){5-7}\cmidrule(lr){8-10}
Model & intact & row off & col off & intact & row off & col off & intact & row off & col off \\
\midrule
naver ViT-B    & 0.469 & 0.159 & 0.234 & 65.5 & \textbf{7.8} & 71.7 & 62.6 & 63.4 & \textbf{7.8} \\
reg1-gap ViT-B & 0.689 & 0.262 & 0.447 & 69.7 & \textbf{7.8} & 74.2 & 69.2 & 69.8 & \textbf{7.6} \\
DINOv3 ViT-B   & 0.949 & 0.370 & 0.649 & 75.4 & \textbf{8.2} & 81.7 & 77.2 & 70.3 & \textbf{8.3} \\
naver ViT-S    & 0.371 & 0.260 & 0.036 & 36.0 & \textbf{7.8} & 31.0 & 50.2 & 49.7 & \textbf{7.9} \\
\bottomrule
\end{tabular}
\end{table}

A random set of 16 planes, one draw on naver ViT-B and DINOv3, sends neither axis to chance: it leaves the row and column probes at 40.5\% and 42.0\% on naver and 41.2\% and 85.0\% on DINOv3. On naver it removes more SSDC than either axis group (0.050, against 0.159 and 0.234) while removing less axis information, so a drop in SSDC does not by itself measure how much positional information is gone. The random planes are matched to the axis groups in number, not frequency.

On DINOv3, removing the row planes also costs the column probe 7 points, and inside a 0--5 window the column probe falls from 83.9\% to 64.7\% at block 6 and from 78.7\% to 65.1\% at block 9 (two of the five failed implementation checks, Appendix~\ref{app:protocol}). This cost does not grow with the window: with row planes off over blocks 0--2, 0--5 and 0--8, the column probe stays at 78.0\%, 79.5\% and 80.7\% while the row probe falls from 65.2\% to 45.6\% to 29.8\%. It is a property of how DINOv3's planes divide between the axes, not of recovery.

\FloatBarrier
\subsection{Absolute position: the border and the prefix tokens}
\label{app:border}

\begin{figure}[!htbp]
\centering
\includegraphics[width=\textwidth]{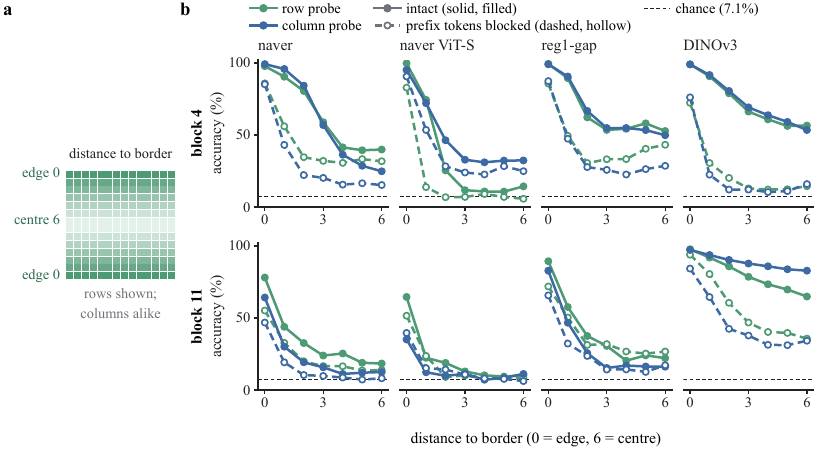}
\caption{\textbf{RoPE models read absolute position from the image border.} \textbf{(a)} Distance to the nearest border along an axis, shown for rows (0: edge, 6: centre); columns are analogous. \textbf{(b)} Row (green) and column (blue) probe accuracy against that distance, at block 4 (top) and block 11 (bottom). Solid, filled: intact; dashed, hollow: attention from patches to the class and register tokens blocked in every block. Dashed black: chance (7.1\%). 1{,}000 images.}
\label{fig:app_border}
\end{figure}

Rotations make attention depend only on the offset between two patches, yet probes read each patch's absolute row and column (Table~\ref{tab:axis}). Two sources could supply it: the image border, where a patch has neighbours on one side only, and the prefix tokens (the class token and, on DINOv3, four registers), which receive no rotation. We test both on the four RoPE models (Figure~\ref{fig:app_border}, Table~\ref{tab:border}). First, we split the row and column probes' test accuracy by the patch's distance to the nearest border along that axis. Second, we block attention from every patch to the prefix tokens in every block, by adding $-\infty$ to those attention logits, and repeat the probes, SSDC and top-1. Probes use a 60/20/20 split with the epoch chosen on validation, 1{,}000 images.

Absolute position is anchored at the border. At block 4, patches on the edge row or column are classified correctly 95--100\% of the time, against 14--56\% at the centre, and accuracy falls steadily with distance from the edge. By block 11 the border advantage remains on every model. Blocking the prefix tokens does not remove this pattern: edge accuracy stays at 72--91\% at block 4 while the centre falls further.

The prefix tokens still contribute. Blocking them lowers block-4 SSDC by a third on naver ViT-S and by four-fifths on DINOv3, whose four registers carry most of its early structure, and costs 1.8, 8.0 and 9.5 top-1 points on naver ViT-B, naver ViT-S and reg1-gap. By block 11, SSDC returns to its intact level on the three single-prefix models (0.38--0.46 against 0.36--0.43 intact) but not on DINOv3 (0.54 against 0.85). The border anchors absolute position; the prefix tokens add to it, mostly in the early blocks.

\begin{table}[!htbp]
\centering
\footnotesize
\setlength{\tabcolsep}{4pt}
\caption{\textbf{Blocking attention to the prefix tokens.} SSDC at blocks 4 and 11, exact-position probe at block 4 (\%), and top-1 (\%). DINOv3 has no classification head.}
\label{tab:border}
\begin{tabular}{lcccccccc}
\toprule
& \multicolumn{2}{c}{SSDC, block 4} & \multicolumn{2}{c}{SSDC, block 11} & \multicolumn{2}{c}{Exact probe, block 4} & \multicolumn{2}{c}{Top-1} \\
\cmidrule(lr){2-3}\cmidrule(lr){4-5}\cmidrule(lr){6-7}\cmidrule(lr){8-9}
Model & intact & blocked & intact & blocked & intact & blocked & intact & blocked \\
\midrule
naver ViT-B    & 0.47 & 0.26 & 0.40 & 0.38 & 28.4 & 11.5 & 84.6 & 82.8 \\
naver ViT-S    & 0.37 & 0.25 & 0.36 & 0.42 & 15.6 & 7.2  & 81.3 & 73.3 \\
reg1-gap ViT-B & 0.69 & 0.40 & 0.43 & 0.46 & 31.6 & 13.5 & 83.9 & 74.4 \\
DINOv3 ViT-B   & 0.95 & 0.19 & 0.85 & 0.54 & 31.9 & 5.5  & --- & --- \\
\bottomrule
\end{tabular}
\end{table}

\FloatBarrier
\subsection{Masking windows}

\begin{figure}[!htbp]
\centering
\includegraphics[width=\textwidth]{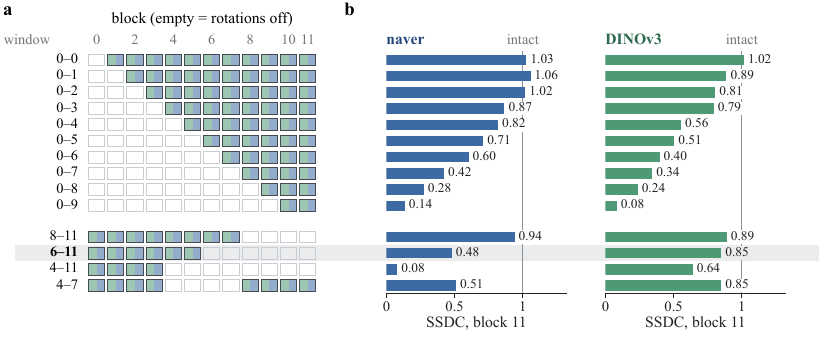}
\caption{\textbf{Which blocks need their rotations.} \textbf{(a)} Each row is one masking window; empty cells are blocks with rotations off, and filled cells show the row (green) and column (blue) rotations that stay on. \textbf{(b)} Normalised SSDC at block 11 (0: floor, 1: intact) for naver ViT-B (left) and DINOv3 (right). naver ViT-B depends on its later blocks (removing 6--11 costs more than 0--5); DINOv3 on its earlier ones.}
\label{fig:app_windows}
\end{figure}

Figure~\ref{fig:app_windows} shows every window in Section~\ref{sec:windows} and how much structure returns by block 11; Table~\ref{tab:windows} gives the values. A prefix window $0$--$k$ switches rotations off from block 0 through block $k$ and restores them afterwards; a suffix window $k$--$11$ switches them off in the last blocks only; a middle window switches them off in an interior range. SSDC at block 11 is read on the input to block 11, so it reflects rotations in blocks 0--10 only; the suffix window 8--11 therefore removes three blocks' rotations, not four.

\begin{table}[!htbp]
\centering
\small
\caption{\textbf{Normalised SSDC at block 11 under each masking window} (0: floor, 1: intact).}
\label{tab:windows}
\begin{tabular}{lcc@{\hskip 18pt}lcc}
\toprule
Rotations off & naver ViT-B & DINOv3 ViT-B & Rotations off & naver ViT-B & DINOv3 ViT-B \\
\midrule
0--0 & 1.03 & 1.02 & 0--6 & 0.60 & 0.40 \\
0--1 & 1.06 & 0.89 & 0--7 & 0.42 & 0.34 \\
0--2 & 1.02 & 0.81 & 0--8 & 0.28 & 0.24 \\
0--3 & 0.87 & 0.79 & 0--9 & 0.13 & 0.08 \\
0--4 & 0.82 & 0.56 & \emph{suffix and middle} & & \\
0--5 & 0.71 & 0.50 & 8--11 & 0.94 & 0.89 \\
     &      &      & 6--11 & 0.48 & 0.85 \\
     &      &      & 4--11 & 0.08 & 0.64 \\
     &      &      & 4--7  & 0.51 & 0.85 \\
\bottomrule
\end{tabular}
\end{table}

For windows that remove one axis, Table~\ref{tab:axis4} gives probe retention at block 11 for the 0--5 window on all four models (the values in Figure~\ref{fig:rope}c), and Table~\ref{tab:axiswin} varies the window length on naver ViT-B and DINOv3. The two tables use different floors: Table~\ref{tab:axis4} uses the block-4 floor, which is at chance, because that run measured the floor only at block 4, while Table~\ref{tab:axiswin} uses the floor at block 11.

\begin{table}[!htbp]
\centering
\small
\caption{\textbf{One axis off in blocks 0--5, then restored.} Probe retention at block 11 (0: floor, 1: intact) for the removed axis and the other axis. Bold: the two models where the other axis ends clearly below intact.}
\label{tab:axis4}
\begin{tabular}{lcccc}
\toprule
& \multicolumn{2}{c}{Row planes off} & \multicolumn{2}{c}{Column planes off} \\
\cmidrule(lr){2-3}\cmidrule(lr){4-5}
Model & row (removed) & column (other) & column (removed) & row (other) \\
\midrule
naver ViT-B    & 0.47 & 1.04 & 0.22 & 1.08 \\
reg1-gap ViT-B & 0.30 & 1.04 & 0.24 & 1.13 \\
DINOv3 ViT-B   & 0.45 & \textbf{0.79} & 0.43 & 1.05 \\
naver ViT-S    & 0.44 & 1.17 & 0.21 & \textbf{0.80} \\
\bottomrule
\end{tabular}
\end{table}

\begin{table}[!htbp]
\centering
\small
\caption{\textbf{Longer windows leave less of the removed axis.} Probe retention at block 11 (0: floor at block 11, 1: intact). Inside the window the removed axis is at the floor in every case.}
\label{tab:axiswin}
\begin{tabular}{llcc}
\toprule
Model & Planes off & Row probe & Column probe \\
\midrule
\multirow{4}{*}{naver ViT-B}
 & row, blocks 0--2    & 0.75 & 0.91 \\
 & row, blocks 0--5    & 0.47 & 1.05 \\
 & row, blocks 0--8    & 0.13 & 1.15 \\
 & column, blocks 0--5 & 1.06 & 0.24 \\
\midrule
\multirow{4}{*}{DINOv3 ViT-B}
 & row, blocks 0--2    & 0.65 & 0.78 \\
 & row, blocks 0--5    & 0.46 & 0.80 \\
 & row, blocks 0--8    & 0.30 & 0.81 \\
 & column, blocks 0--5 & 1.05 & 0.43 \\
\bottomrule
\end{tabular}
\end{table}

% ---------------------------------------------------------------------------
\FloatBarrier
\section{Head attribution}
\label{app:staged}

This appendix supports Section~\ref{sec:heads}. Throughout, a ``head'' is one head index across the blocks named.

\FloatBarrier
\subsection{Axis specialisation}

Table~\ref{tab:staged} gives the numbers behind Figure~\ref{fig:rope}d. Each entry is the probe points lost when one head's rotations stay off in blocks 6--11 during recovery from a one-axis 0--5 window, relative to the same window with every head restored. Because SSDC and probes are read on the input to each block, a head whose rotations are removed from block 6 first affects the reading at block 7, so the cost is averaged over blocks 7--11. The row cost uses the row-removed window and the column cost the column-removed window. The heads are the top four by the SSDC rebuilding score (three on naver ViT-S), and the three-standard-deviation noise bar, measured by repeating the probe, is about 1.2 points.

\begin{table}[!htbp]
\centering
\footnotesize
\caption{\textbf{Cost of keeping one head's rotations off during recovery}, in probe points (row window / column window), averaged over blocks 7--11. $\dagger$: below the 1.2-point noise bar.}
\label{tab:staged}
\begin{tabular}{lcccc}
\toprule
Model & \multicolumn{4}{c}{Head: row cost / column cost} \\
\midrule
naver ViT-B    & h6: 5.19 / 0.09 & h1: 3.78 / 0.25 & h2: 1.86 / 0.18 & h11: 0.48 / 0.08 \\
DINOv3 ViT-B   & h2: 0.93 / 9.26 & h9: 0.99 / 5.20 & h1: 3.83 / 3.39 & h10: 1.85 / 2.76 \\
naver ViT-S    & h0: 3.03 / 0.78 & h4: 2.02 / 0.28 & h2: $0.73^\dagger$ / $0.68^\dagger$ & --- \\
reg1-gap ViT-B & h7: 3.84 / 0.00 & h9: 3.08 / 0.19 & h3: 0.30 / 2.70 & h6: 0.05 / 0.00 \\
\bottomrule
\end{tabular}
\end{table}

On naver ViT-S, head 2, the top rebuilder by SSDC, costs neither axis more than the noise bar, yet it is the costliest head for classification (Appendix~\ref{app:cls_rope}). Axis specialisation is therefore clear on naver ViT-B, DINOv3 and reg1-gap, and absent for the top head on naver ViT-S.

\FloatBarrier
\subsection{Carrying versus rebuilding}
\label{apd:carry}

We also scored each head's \emph{carrying} importance: the normalised SSDC lost at block 4 when that head index's rotations are off in every block, during ordinary inference. Table~\ref{tab:heads} compares it with the rebuilding score. The two cannot be compared directly: a head index shares no parameters across blocks, and the carrying score depends only on blocks 0--3 while the rebuilding score depends only on blocks 6--11, so they rank different sets of weights. Reading carrying at block 11 instead changes the ranking: on naver ViT-B the top rebuilder, head 6, becomes the second most important carrier, and the rank correlation between carrying and rebuilding rises from $+0.06$ to $+0.88$; on DINOv3 the top rebuilder, head 9, stays the least important carrier ($-0.17$ at block 4, $+0.27$ at block 11). The rebuilding scores are also close together: on DINOv3 the top three heads are within 0.004 of each other.

\begin{table}[!htbp]
\centering
\small
\caption{\textbf{Carrying and rebuilding rankings.} ``Concentration'': share of SSDC lost at block 4 when the single most important carrying head is removed during ordinary inference. ``Rank as carrier'': where the top rebuilder sits in the carrying ranking (1 = most important). $\rho$: Spearman correlation of the two per-head rankings.}
\label{tab:heads}
\begin{tabular}{lccccc}
\toprule
Model & Heads & Concentration & Top rebuilder & Rank as carrier & $\rho$ \\
\midrule
naver ViT-B    & 12 & 31\% (head 2)  & head 6 & 11 of 12 & $+0.06$ \\
DINOv3 ViT-B   & 12 & 88\% (head 2)  & head 9 & 11 of 12 & $-0.18$ \\
naver ViT-S    &  6 & 71\% (head 4)  & head 2 &  5 of 6  & $-0.09$ \\
reg1-gap ViT-B & 12 & 46\% (head 0)  & head 7 &  7 of 12 & $-0.35$ \\
\bottomrule
\end{tabular}
\end{table}

Concentration is not set by head count or scale: the three 12-head models span 31--88\%, and the two naver checkpoints, which share a recipe, differ (31\% and 71\%).

\FloatBarrier
\subsection{Attention statistics}
\label{app:attn}

One might expect rebuilding heads to look further across the image than carrying heads, but they do the opposite: over blocks 6--11 on 64 images, the top two rebuilders attend nearer and with lower entropy than the top two carriers on all four models (Table~\ref{tab:attn}), after excluding any head that appears in both lists. The bracketed figure is the difference as a share of the largest head-to-head difference in that model. Each group holds one or two heads, so we report the direction, not a tested difference, and these statistics do not explain what the rebuilding heads compute.

\begin{table}[!htbp]
\centering
\small
\caption{\textbf{Top two rebuilders minus top two carriers}, blocks 6--11. Negative: the rebuilding heads attend nearer, or with lower entropy.}
\label{tab:attn}
\begin{tabular}{lcccc}
\toprule
Statistic & naver ViT-B & DINOv3 ViT-B & naver ViT-S & reg1-gap ViT-B \\
\midrule
Mean distance (patches) & $-0.05$ (4\%) & $-0.87$ (41\%) & $-0.19$ (18\%) & $-0.64$ (80\%) \\
Entropy (nats)          & $-0.03$ (6\%) & $-0.30$ (44\%) & $-0.08$ (26\%) & $-0.11$ (17\%) \\
Attention to prefix     & $+0.00$       & $+0.02$        & $+0.02$        & $-0.02$ \\
\bottomrule
\end{tabular}
\end{table}

% ---------------------------------------------------------------------------
\FloatBarrier
\section{Comparison: a shared recipe and a shared model}
\label{app:compare}

This appendix supports Section~\ref{sec:compare}.

\FloatBarrier
\subsection{DeiT-III and naver ViT-B}
\label{app:deit}

DeiT-III ViT-B (APE) and naver ViT-B (RoPE) are trained with the same recipe and reach the same top-1 on our 5{,}000 evaluation images (84.58\% and 84.64\%), so their comparison is not confounded by training. DeiT-III's SSDC is 0.57 entering block 0, rises to 0.84 at blocks 5--6 and falls to 0.38 at block 11; naver's starts at 0.03 and reaches 0.46 (Figure~\ref{fig:depth}). Zeroing DeiT-III's embedding leaves SSDC at the floor (at most 0.011) at every block and top-1 at 23.5\%, so, like the other APE models, it does not rebuild position once its only source is gone. Permuting the embedding across patches instead keeps exact position decodable (99.7\% at block 2) but removes SSDC and costs 52 top-1 points.

\FloatBarrier
\subsection{The hybrid checkpoint}
\label{app:hybrid}

\begin{figure}[!htbp]
\centering
\includegraphics[width=0.5\textwidth]{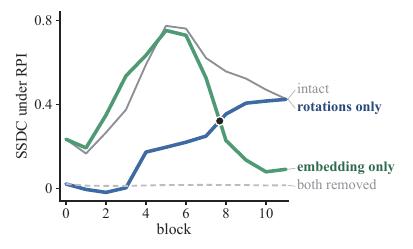}
\caption{\textbf{In the hybrid, the embedding builds early structure and the rotations late structure.} SSDC under RPI entering each block of the hybrid ViT-B: intact, rotations off, embedding removed, and both removed. 1{,}000 images.}
\label{fig:app_hybrid}
\end{figure}

The hybrid ViT-B has both an added embedding and rotations. With every rotation removed, SSDC exceeds the intact curve in blocks 1--4, stays at 96--97\% of it through block 6, and then collapses, reaching 0.228 at block 8 against 0.557 intact (Figure~\ref{fig:app_hybrid}). With the embedding removed instead, SSDC starts near zero and climbs with depth, passing the rotations-off curve at block 8 (0.354 against 0.228). The embedding supplies the hybrid's early structure and the rotations its late structure, the same depth signatures as the separate APE and RoPE models.

Position does not need the rotations to be readable, but the classifier needs them. At block 4, the exact-position probe reads 85.1\% intact, 88.6\% with every rotation off, 4.1\% with the embedding removed, and 0.8\% with both removed (chance 0.5\%). Top-1 is 80.5\% intact and 37.4\% with the rotations off. On this checkpoint the top rebuilder costs no more accuracy than ordinary heads, so it is not in Figure~\ref{fig:accuracy}b, and we did not run the all-heads test on it. All hybrid values are raw: with the rotations off the hybrid stays above its intact SSDC in early blocks, so the floor used for normalisation elsewhere does not exist here.

% ---------------------------------------------------------------------------
\FloatBarrier
\section{Classification}
\label{app:cls}

This appendix supports Section~\ref{sec:accuracy}: every APE arm, every RoPE head, a replication from an earlier run, and why that section reports accuracy rather than SSDC.

\FloatBarrier
\subsection{APE: every edit}
\label{app:cls_ape}
\label{apd:ape_functional}

Table~\ref{tab:cls_ape} gives top-1 for every APE edit. Supervised ViT-B and DeiT-III use 5{,}000 images, AugReg and SAM 1{,}000 (binomial standard error about 1.3 points); DINO has no classification head. On supervised ViT-B, paired over the same 5{,}000 images, swapping the features to other patches costs $24.8 \pm 0.7$ points more than swapping them to the same patch of another image, and zeroing them costs $15.1 \pm 0.7$ more than zeroing a matched random set, both with the error added back. On SAM the same two gaps are $31.4 \pm 1.5$ and $13.9 \pm 1.3$ points over 1{,}000 images; per-image results for AugReg were not kept, so we give no paired errors there.

\begin{table}[!htbp]
\centering
\footnotesize
\caption{\textbf{Top-1 accuracy (\%) under each APE edit.} Edits act at the residual stream entering block 2, except the embedding rows. Matched random: mean $\pm$ standard deviation over five draws.}
\label{tab:cls_ape}
\begin{tabular}{lcccc}
\toprule
Edit & supervised & AugReg & SAM & DeiT-III \\
\midrule
Intact & 81.16 & 78.6 & 81.0 & 84.58 \\
SAE reconstruction & 79.34 & 78.3 & 80.5 & --- \\
Positional zeroed & 50.32 & 50.9 & 54.8 & --- \\
Uniform random zeroed & 78.94 & 77.4 & 80.5 & --- \\
Matched random zeroed & 73.90 $\pm$ 0.75 & 73.8 $\pm$ 0.4 & 78.8 $\pm$ 0.6 & --- \\
\midrule
\multicolumn{5}{l}{\emph{With the SAE error added back}} \\
Positional zeroed & 61.52 & 56.6 & 65.2 & --- \\
Positional mean-ablated & 62.78 & 59.8 & 65.1 & --- \\
Swapped, same patch of another image & 80.84 & 77.9 & 80.7 & --- \\
Swapped, other patch of the same image & 56.04 & 53.7 & 49.3 & --- \\
Matched random zeroed (one draw) & 76.60 & 75.3 & 79.1 & --- \\
\midrule
Embedding zeroed & 24.88 & 28.3 & 40.0 & 23.52 \\
Embedding permuted & 28.50 & 29.7 & 23.1 & 32.12 \\
\bottomrule
\end{tabular}
\end{table}

The matched random features cost 2--5 points beyond the reconstruction, far more than the uniformly drawn ones, yet leave spatial structure intact (Table~\ref{tab:ape_controls}), so they measure what removing active but position-free features costs. The same-patch swap costs at most 0.7 points while the other-patch swap costs 25--32, so the classifier depends on which position the features report.

\begin{figure}[!htbp]
\centering
\includegraphics[width=\textwidth]{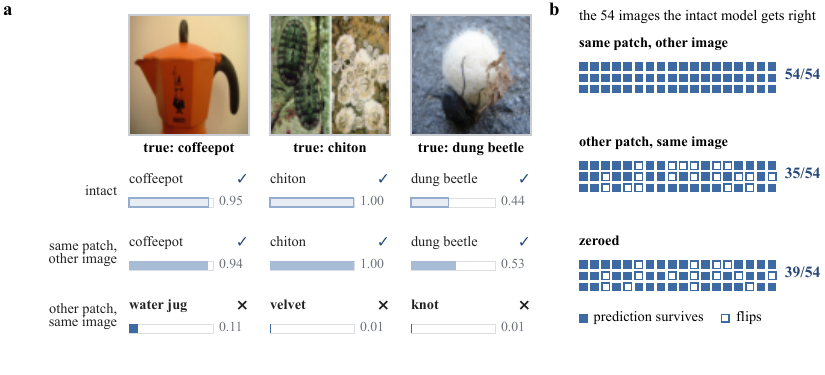}
\caption{\textbf{Moving the positional features changes the prediction, although the image does not change.} Supervised ViT-B, with its 20 positional features edited at block 2 and the SAE error added back. \textbf{(a)} Top-1 prediction and the probability of the true class, intact, with the features swapped with the same patch of another image, and with them swapped between patches of the same image. The examples are the first three validation images that the intact model classifies correctly and the other-patch swap does not, skipping images that show people. \textbf{(b)} The 54 of the first 64 validation images that the intact model classifies correctly: filled if the prediction survives the edit.}
\label{fig:app_real_predictions}
\end{figure}

Figure~\ref{fig:app_real_predictions} shows the same effect on single images. The pixels are untouched and only the 20 positional features are edited. Moving them to the same patch of another image leaves every prediction intact, while moving them between patches of the same image turns a coffee pot into a water jug and a chiton into velvet. Of the first 64 validation images, the intact model classifies 54 correctly; the same-patch swap keeps all 54, the other-patch swap 35 and zeroing 39.

\FloatBarrier
\subsection{RoPE: every head}
\label{app:cls_rope}

\begin{table}[!htbp]
\centering
\footnotesize
\setlength{\tabcolsep}{4pt}
\caption{\textbf{Top-1 cost (points) of removing one head's rotations in blocks 6--11.} Bold: the top rebuilder, fixed in advance by its SSDC score. Reference rows give top-1 (\%). ``Shuffled'' replaces rotations with those of a fixed random permutation of patch positions instead of the identity (two seeds for all blocks).}
\label{tab:cls_rope}
\begin{tabular}{lcc@{\hskip 12pt}cc@{\hskip 12pt}cc}
\toprule
& \multicolumn{2}{c}{naver ViT-B} & \multicolumn{2}{c}{reg1-gap ViT-B} & \multicolumn{2}{c}{naver ViT-S} \\
\cmidrule(lr){2-3}\cmidrule(lr){4-5}\cmidrule(lr){6-7}
Head & window & intact & window & intact & window & intact \\
\midrule
0  & 1.34 & 0.12 & 0.70 & 0.20 & 3.70 & 0.50 \\
1  & 3.38 & 0.44 & 0.24 & 0.24 & 1.00 & 0.96 \\
2  & 1.64 & 0.30 & 3.94 & 0.22 & \textbf{7.62} & \textbf{1.18} \\
3  & $-$0.14 & 0.52 & 4.16 & 0.28 & 4.88 & 1.40 \\
4  & 2.18 & 0.54 & $-$1.56 & 0.18 & 6.48 & 1.08 \\
5  & 2.36 & 0.14 & 0.98 & 0.16 & 0.34 & 1.78 \\
6  & \textbf{4.30} & \textbf{0.50} & 1.52 & 0.24 & & \\
7  & $-$0.10 & 0.12 & \textbf{5.74} & \textbf{0.42} & & \\
8  & 0.36 & 0.36 & $-$0.92 & $-$0.04 & & \\
9  & 1.82 & 0.42 & 4.90 & 0.20 & & \\
10 & $-$1.66 & 0.36 & $-$1.24 & 0.78 & & \\
11 & 2.30 & 0.44 & $-$0.46 & 0.10 & & \\
\midrule
Top rebuilder, shuffled & 3.52 & 0.42 & 5.86 & 0.40 & 6.22 & 1.02 \\
\midrule
\multicolumn{7}{l}{\emph{Reference top-1 (\%)}} \\
Intact & \multicolumn{2}{c}{84.64} & \multicolumn{2}{c}{83.88} & \multicolumn{2}{c}{81.32} \\
Window (0--5 off) & \multicolumn{2}{c}{70.52} & \multicolumn{2}{c}{56.28} & \multicolumn{2}{c}{48.68} \\
Window, shuffled & \multicolumn{2}{c}{67.16} & \multicolumn{2}{c}{57.36} & \multicolumn{2}{c}{53.38} \\
All rotations off & \multicolumn{2}{c}{38.84} & \multicolumn{2}{c}{29.50} & \multicolumn{2}{c}{24.04} \\
All shuffled (two seeds) & \multicolumn{2}{c}{34.20 / 33.40} & \multicolumn{2}{c}{22.58 / 22.74} & \multicolumn{2}{c}{20.78 / 20.72} \\
\bottomrule
\end{tabular}
\end{table}
\label{app:acc}

Table~\ref{tab:cls_rope} gives, for every head, the top-1 cost of removing its rotations in blocks 6--11, either after the 0--5 window (``window'') or in the unmodified model (``intact''). All runs use 5{,}000 images at 224$\times$224; paired standard errors of the window costs are 0.20--0.36 points on naver ViT-B and 0.37--0.46 on naver ViT-S. Per-image results for reg1-gap were overwritten during a rerun, so it has no paired errors; its heads' window costs have a standard deviation of 2.6 points.

The top rebuilder is the costliest head in the window on all three models, and in the intact model every head costs at most 0.8 points on the two ViT-B models and 1.8 on naver ViT-S. Replacing the rotations with shuffled ones costs about as much as removing them, both for all heads and for the top rebuilder alone, so the costs come from missing position rather than from attention receiving rotations it never saw in training.

\FloatBarrier
\subsection{Replication and higher resolution}
\label{apd:ssdc_highres}

An earlier run of the window test, on a different set of 5{,}000 images and with three random heads per model, gives the same top-rebuilder costs: 4.40, 5.82 and 8.76 points on naver ViT-B, reg1-gap and naver ViT-S, against 4.30, 5.74 and 7.62 here. Its absolute accuracies are 2--6 points lower on every arm because it used a 256$\times$256 repack of the validation set (Appendix~\ref{app:protocol}). That run also covered 320$\times$320 input, where the top rebuilder costs 7.84, 6.08 and 7.80 points after the window.

At 320$\times$320, SSDC loses sensitivity. Baseline SSDC at the last block falls by about a third on the three models trained at 224 while top-1 barely moves, and by 4\% on reg1-gap, which was trained at 256. On naver ViT-B at 320, a 1{,}000-image run with the top rebuilder (head 6) and a control head (head 0) gives top-1 losses of 6.95 and 3.13 points, while their SSDC costs are 0.005 and 0.003, a gap of 0.002. SSDC does not separate them; accuracy does. reg1-gap and naver ViT-S rank their heads the same way under both measures at both resolutions, so this is SSDC losing sensitivity in one setting rather than a general dissociation, and it is why Section~\ref{sec:accuracy} uses accuracy.

% ---------------------------------------------------------------------------
\FloatBarrier
\section{Reproducibility: code, data, compute and protocol}
\label{app:protocol}

\label{app:code}
\paragraph{Code.} The code, with a script per experiment and the result files (JSON with per-block SSDC, probe accuracies and top-1, plus per-image correctness for the accuracy runs) from which every table and figure is generated, is available at \url{https://github.com/mahmoud-mannes/vit-sae-analysis}.

\paragraph{Models and data.} Every checkpoint is public (Table~\ref{tab:models}). Images are the ImageNet-1k validation split, streamed from the HuggingFace hub. The rotation-masking runs (Section~\ref{sec:rope_mech}; Appendices~\ref{app:windows}--\ref{app:staged}) use a 256$\times$256 repack (\texttt{benjamin-paine/imagenet-1k-256x256}); the control runs and the patching experiments use the original images (\texttt{ILSVRC/imagenet-1k}). Each experiment takes the first $N$ images of the stream in a fixed order, with fixed seeds for RPI permutations and random feature draws. The SAE weights and the lists of selected positional features will be released with the code.

\paragraph{Compute.}\label{app:compute} The experiments ran on four machines (Table~\ref{tab:compute}). Everything runs in float32 with \texttt{timm} 1.0.29; PyTorch 2.10 for the earlier runs and 2.14 for the control runs.

\begin{table}[!htbp]
\centering
\footnotesize
\caption{\textbf{Hardware per group of experiments.}}
\label{tab:compute}
\begin{tabular}{@{}>{\raggedright\arraybackslash}p{8.2cm}>{\raggedright\arraybackslash}p{4.2cm}@{}}
\toprule
Experiments & Hardware \\
\midrule
SAE training, feature screen, APE dose-response, axis and depth probes (\S\ref{sec:ape_mech}; App.~\ref{apd:sae_config}--\ref{apd:ape_depth_probes}, \ref{apd:ape_axis_crossmodel}) & 1$\times$ NVIDIA T4 (16 GB) \\
Patching, zero-ablation and alignment controls (App.~\ref{sec:results_alignment}) & 1$\times$ NVIDIA RTX A4000 (16 GB) \\
Rotation masking, windows, heads, first accuracy run (\S\ref{sec:rope_mech}; App.~\ref{app:windows}, \ref{app:staged}) & 1$\times$ NVIDIA H100 (80 GB) \\
Control runs: APE controls, border, DeiT-III, all-heads accuracy (App.~\ref{apd:ape_controls}, \ref{app:border}, \ref{app:deit}, \ref{app:cls}) & 2$\times$ NVIDIA H100 (80 GB) \\
\bottomrule
\end{tabular}
\end{table}

\paragraph{Sample sizes.} SSDC and probes use 1{,}000 ImageNet-1k validation images unless stated (10{,}000 for the supervised ViT-B dose-response). The patching experiments use 512 images with two seeds. Top-1 uses 5{,}000 images for supervised ViT-B, DeiT-III and every RoPE model, and 1{,}000 for AugReg and SAM, whose binomial standard error is about 1.3 points.

\paragraph{Probes.} Linear probes use AdamW \citep{loshchilov2019decoupled} at learning rate $5\times10^{-3}$ for 20 passes with batch size 512, on standardised features, with the same settings for every arm. The earlier runs split images 80/20 and report the best test accuracy over epochs. The control runs split images 60/20/20 and choose the epoch on the validation split. Across the more than 400 probes in the control runs, the old protocol would have reported 0.1 points more on average and at most 0.7, so it does not change any comparison. The APE axis probes are ridge regressions on the normalised row and column coordinates, split 600/200/200 by image.

\paragraph{Floors and normalisation.} On the four RoPE-only models the floor sits at 0.008--0.029 SSDC at every block, against intact values of 0.37--0.95. Block 0 comes before any rotation, so its intact value equals its floor and normalised values are undefined there. The APE floor zeroes the added embedding (SSDC at most 0.021 at any block). Hybrid results stay raw (Appendix~\ref{app:hybrid}).

\paragraph{Rotation masking.} Each ViT-B RoPE head pairs its 64 query and key dimensions into 32 rotation planes, 16 per axis, assigned to axes at runtime from each model's frequency table. Setting a plane to the identity (sine 0, cosine 1 in the model's cached tables) is exact and reversible and can target one axis, block range or head. Per-head masking uses the fact that the cached rotation tensor has no head dimension: broadcasting a per-head tensor in its place applies the identity to the chosen heads only. We checked this against the source code of both attention classes these checkpoints use.

\paragraph{Implementation checks.} Every experiment runs a set of checks before its main loop: that the planes partition correctly, that hooks fire the expected number of times, that readouts sit at the intended block, and that a no-op mask gives a bit-identical forward pass. Of 318 checks in the earlier runs, 313 passed. The five failures are reported where they apply: three record that the hybrid's rotations-off SSDC sits above its intact value (Appendix~\ref{app:hybrid}), which comes from its second positional signal, and two record DINOv3's cross-axis cost (Appendix~\ref{app:windows}). All 15 checks passed in each of the three control runs on RoPE models. On the APE side, adding the SAE error back with no edit reproduces the intact model's SSDC and top-1 exactly.

\paragraph{Missing per-image results.} Paired standard errors use per-image correctness saved for every arm. For reg1-gap, a rerun after a machine restart overwrote the per-image files of the earlier arms, so only their means are available; we give the spread across heads instead.

\end{document}